\documentclass[10pt,twocolumn,letterpaper]{article}

\usepackage[pagenumbers]{cvpr} % To force page numbers, e.g. for an arXiv version

\usepackage[table]{xcolor}
\usepackage{arydshln}           % 表格虚线
\usepackage{caption}
\usepackage{lipsum}
\usepackage{algorithm,algorithmic}
\usepackage{multirow}
\usepackage{setspace}
\usepackage{ulem}
\usepackage{placeins}
\usepackage{makecell}
\definecolor{cvprblue}{rgb}{0.21,0.49,0.74}
\usepackage[pagebackref,breaklinks,colorlinks,citecolor=cvprblue]{hyperref}

\def\paperID{809} % *** Enter the Paper ID here
\def\confName{CVPR}
\def\confYear{2024}

\title{BEVSpread: Spread Voxel Pooling for Bird’s-Eye-View Representation \\ in Vision-based Roadside 3D Object Detection}

\newcommand*\samethanks[1][\value{footnote}]{\footnotemark[#1]}

\author{
Wenjie Wang$^1$\thanks{The first two authors contributed equally to this paper.}, 
Yehao Lu$^1$\samethanks[1],
Guangcong Zheng$^1$,
Shuigen Zhan$^2$,
Xiaoqing Ye$^3$,
Zichang Tan$^3$\\
Jingdong Wang$^3$,
Gaoang Wang$^{1}$,
Xi Li$^{1,2,4}$\thanks{Corresponding author.}\\
$^1$College of Computer Science and Technology, Zhejiang University \\
$^2$Polytechnic Institute, Zhejiang University \enspace $^3$Baidu \\
$^4$Zhejiang – Singapore Innovation and AI Joint Research Lab\\
{\tt\small \{wenjie\_wang, luyehao, guangcongzheng, shuigenzhan, xilizju\}@zju.edu.cn}\\
{\tt\small gaoangwang@intl.zju.edu.cn}\enspace
{\tt\small \{yexiaoqing, tanzichang, wangjingdong\}@baidu.com}
}

\begin{document}

\twocolumn[{%
\renewcommand\twocolumn[1][]{#1}%
\maketitle
\vspace{-9mm}

\begin{center}
    \centering
    \captionsetup{type=figure}
    \includegraphics[width=1\textwidth]{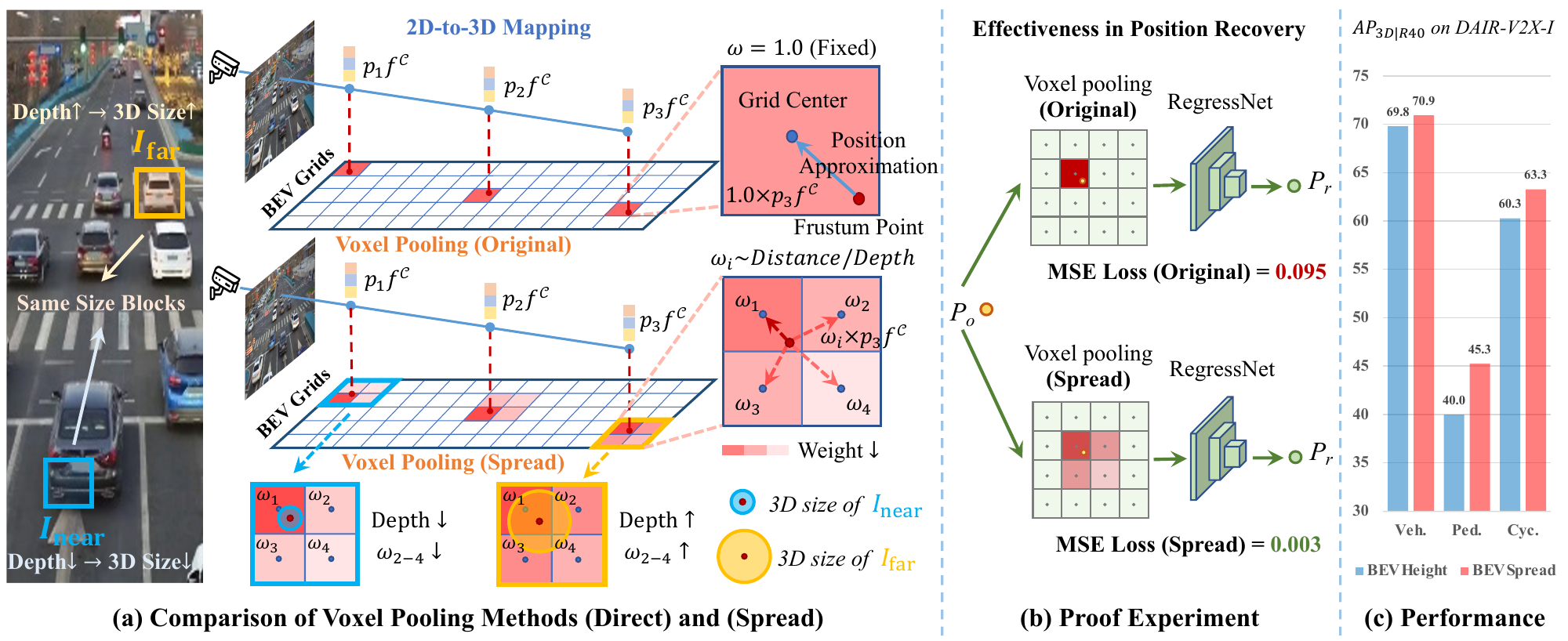}
    \caption{\textbf{(a)} Original voxel pooling strategy approximately accumulates the image features contained in a frustum point to the single corresponding BEV grid center, leading to an irrecoverable position approximation error. We discover that spread operation can reduce this error, where the weights $\omega$ assigned to surrounding BEV grids should be related to the distance and depth. First, weight decay with distance can effectively retain more location information, which is beneficial for subsequent network learning. Second, same size image blocks with deeper depth represent objects of larger 3D scales, which results in distant objects containing few image features. Therefore, it is reasonable to assign larger weights to the surrounding BEV grids for distant targets. \textbf{(b)} We have designed an intuitive experiment to demonstrate that network can learn accurate position coordinates from the BEV features obtained by spread voxel pooling. \textbf{(c)} Results on DAIR-V2X-I dataset show that BEVSpread outperforms state-of-the-art method by a significant margin of (1.12, 5.26, 3.01) AP in vehicle, pedestrian and cyclist categories, respectively.}
    \label{first_img}
\end{center}
}]
\def\thefootnote{*}\footnotetext{Equal contribution.}
\def\thefootnote{$\dagger$}\footnotetext{Corresponding author.}

\begin{abstract}
Vision-based roadside 3D object detection has attracted rising attention in autonomous driving domain, since it encompasses inherent advantages in reducing blind spots and expanding perception range. While previous work mainly focuses on accurately estimating depth or height for 2D-to-3D mapping, ignoring the position approximation error in the voxel pooling process. Inspired by this insight, we propose a novel voxel pooling strategy to reduce such error, dubbed BEVSpread. Specifically, instead of bringing the image features contained in a frustum point to a single BEV grid, BEVSpread considers each frustum point as a source and spreads the image features to the surrounding BEV grids with adaptive weights. To achieve superior propagation performance, a specific weight function is designed to dynamically control the decay speed of the weights according to distance and depth. Aided by customized CUDA parallel acceleration, BEVSpread achieves comparable inference time as the original voxel pooling. Extensive experiments on two large-scale roadside benchmarks demonstrate that, as a plug-in, BEVSpread can significantly improve the performance of existing frustum-based BEV methods by a large margin of (1.12, 5.26, 3.01) AP in vehicle, pedestrian and cyclist. The source code will be made publicly available at  \href{https://github.com/DaTongjie/BEVSpread}{BEVSpread}.
\end{abstract}   
\section{Introduction}
\label{sec:Introduction}
Vision-centric 3D object detection plays a critical role in autonomous driving perception, which helps accurately estimate the state of the surrounding environment and provide reliable observations for forecasting and planning at a low cost. Most existing work focuses on the ego vehicle system \cite{OFT, LSS, BEVFormer, BEVDepth, StreamPETR, FrustumFormer}, facing safety challenges due to a lack of global perspective and the limitation of long-range perception capacity. To this end, roadside 3D object detection has attracted rising attention in recent years \cite{BEVHeight, BEVHeight2, CBR, MonoGAE, AR2VP}. Since roadside cameras are mounted on poles a few meters above the ground, they have inherent advantages in reducing blind spots, improving occlusion robustness, and expanding global perception capability \cite{DAIR-V2X, Rope3D, V2X-Seq, A9-Dataset}. Therefore, it is promising to improve roadside perception performance as a complement to improve the safety of autonomous driving.

Recently, bird’s eye view (BEV) has become the mainstream paradigm for handling the 3D object detection task \cite{vision_BEV_Survey_1, vision_BEV_Survey_2}, among which frustum-based method \cite{LSS, BEVDepth, BEVHeight, SA-BEV} is a significant branch and its pipeline is shown in \cref{first_img}\textcolor{red}{a}. It first maps image features to 3D frustums by estimating depth or height, and then pools frustums onto BEV grids by reducing the Z-axis degree of freedom. Extensive work focuses on improving the precision of depth estimation \cite{LSS, BEVDet, BEVPoolv2, BEVDet4D, BEVDepth} or height estimation \cite{BEVHeight, BEVFormer, HeightFormer} to improve the performance of 2D-to-3D mapping. However, the approximation error caused by the voxel pooling process is rarely considered. As shown in \cref{first_img}\textcolor{red}{a}, the predicted point is usually not located in a BEV grid center. To improve the computational efficiency, previous work approximately accumulates the image features contained in the predicted point to the single corresponding BEV grid center, leading to a position approximation error, and this error is irrecoverable. Augmenting the density of BEV grids can alleviate this error, but results in a notable increase in computational workload. Especially in roadside scenarios, due to the large perception range and limited computing resources, BEV grids can only be designed relatively sparse to ensure real-time detection, which exactly exacerbates the impact of this error. Thus, the question is raised: How can we reduce this error while maintaining computational complexity?

In this work, we propose a novel voxel pooling strategy to reduce such position approximation error, dubbed BEVSpread. Instead of adding the image features contained in a frustum point to a single BEV grid, BEVSpread considers each frustum point as a source and spreads the image features to the surrounding BEV grids with adaptive weights. We discover that the weights assigned to surrounding BEV grids should be related to distance and depth. First, weight decay with distance can effectively retain more location information, which is beneficial for subsequent network learning. Second, we notice that same size image blocks with deeper depth represent objects of larger 3D scales, which results in distant objects containing few image features. Therefore, it is reasonable to assign larger weights to the surrounding BEV grids for distant targets. Inspired by this insight, a specific weight function is designed to achieve superior spread performance, where weights and distances follow a Gaussian distribution. The variance of this Gaussian distribution is positively related to the depth information, which controls the decay speed. In particular, BEVSpread is a plug-in that can be directly deployed on existing frustum-based BEV methods.

To validate the effectiveness of BEVSpread, extensive experiments are conducted on two challenging benchmarks for vision-based roadside perception, DAIR-V2X-I \cite{DAIR-V2X} and Repo3D \cite{Rope3D}. After deploying spread voxel pooling strategy, the 3D average precision ($\text{AP}_{\text{3D}|\text{R40}}$) of BEVHeight \cite{BEVHeight} and BEVDepth \cite{BEVDepth} increases by a large margin of 3.1 and 4.0 on average across three major categories. 

Our contributions can be summarized as:
\begin{itemize}
  \item We point out a position approximation error existed in current voxel pooling approach, which seriously affects the performance of 3D object detection in roadside scenarios, while this issue is ignored in previous works. 
  \item We propose a novel spread voxel pooling approach, namely BEVSpread, which considers both distance and depth effects during the spread process to reduce the position approximation error while maintaining comparable inference time through CUDA parallel acceleration.
  % \item We design a specific weight function to dynamically control the variation of the decay speed, achieving superior spread performance.
  \item Extensive experiments demonstrate that, as a plug-in, BEVSpread significantly enhances the performance of existing frustum-based BEV methods by a large margin of (1.12, 5.26, 3.01) AP in vehicle, pedestrian, and cyclist categories, respectively.
\end{itemize}

\section{Related Work}
\label{sec:RelatedWork}

\begin{figure*}[htbp] %H为当前位置，!htb为忽略美学标准，htbp为浮动图形
    \centering
    \includegraphics[width=1\textwidth]{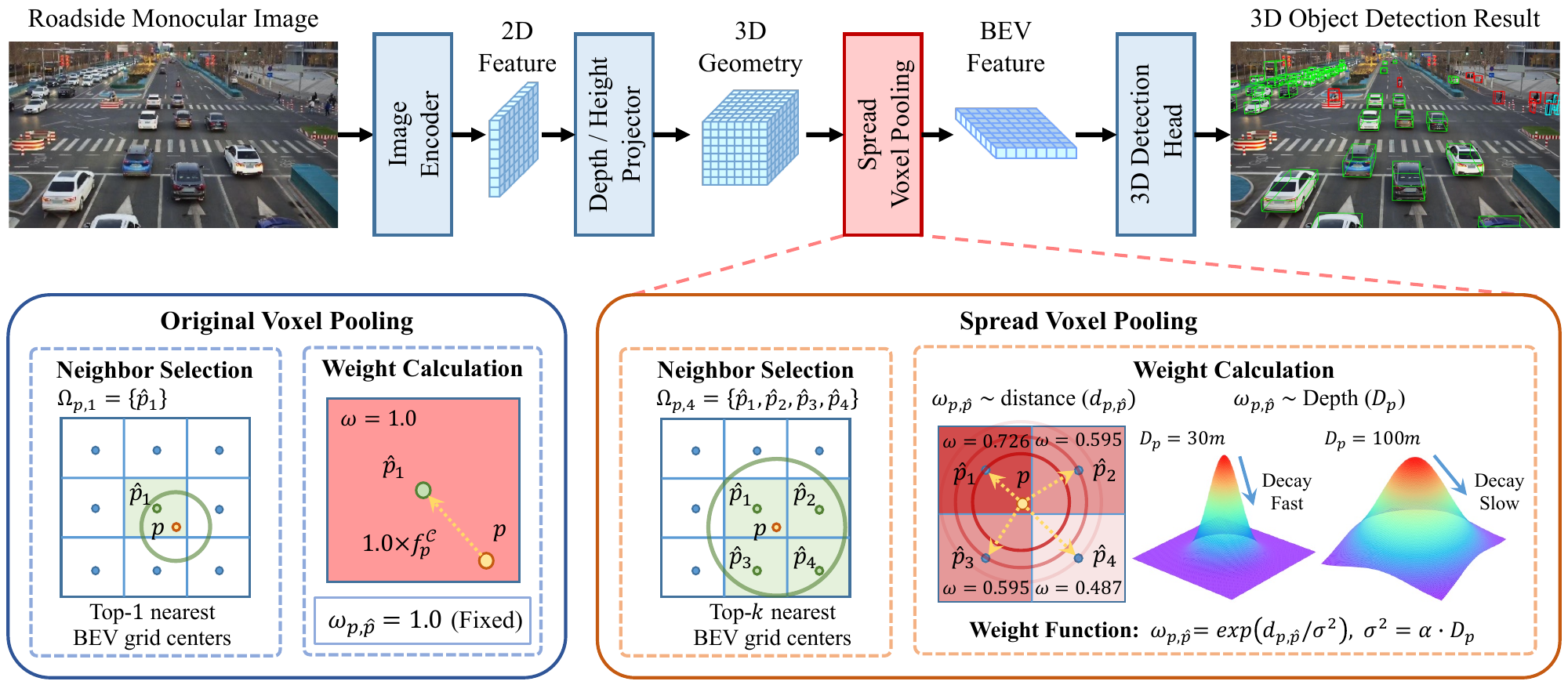}
    % \vspace{-10mm}
    \caption{\textbf{The overall framework of BEVSpread.} Spread voxel pooling consists of two main steps, Neighbor Selection and Weight Calculation. First, each 3D geometry point $p$ is mapped to BEV space, where $top-k$ nearest BEV grid centers are selected as its neighbors $\Omega_{p,k}$. Correspondingly, the original voxel pooling selects the $top-1$ nearest BEV grid center as its neighbor $\Omega_{p,1}$. Second, the weights are calculated for the neighbors by Weight Function, where the weights $\omega_{p,\hat{p}}$ and the distances $d_{p,\hat{p}}$ follow a Gaussian distribution with $(0, \sigma^2)$. Furthermore, the variance $\sigma^2$ is positively related to depth $D_p$, which controls the decay speed of $\omega_{p,\hat{p}}$. Ultimately, the image features contained in each 3D geometry point are accumulated to its neighbors according to the calculated weights.}
    \label{Pipeline}
\end{figure*}

Recently, bird’s eye view (BEV) has become the mainstream paradigm for 3D object detection in autonomous driving, as it provides a unified feature space for multi-sensor and clearly presents the location and scale of objects. In this section, we introduce BEV perception, roadside BEV perception and voxel pooling strategy in detail.

\noindent \textbf{BEV Perception.}
Based on the sensor types, BEV approaches can be mainly divided into three parts including vision-based \cite{OFT, LSS, DETR3D, BEVDepth, BEVFormer, BEVHeight}, LiDAR-based \cite{PointPillars, PV-RCNN, SA-SSD, SST, Afdetv2, DSVT} and fusion-based \cite{Bevfusion, UVTR, UniFusion, SparseFusion, CMT} methods. Benefits from its low cost for deployment, vision-based BEV methods have been a topic of great significance, which are further divided into transformer-based and frustum-based schema. Transformer-based methods \cite{DETR3D, BEVFormer, PETR, PolarFormer, PETRv2, StreamPETR} introduce 3D object queries or BEV grid queries to regress 3D bounding boxes. Frustum-based methods \cite{LSS, CaDDN, BEVDepth, Bevstereo, BEVHeight, SA-BEV} first map image features to 3D frustums by estimating depth or height and then generate BEV features by voxel pooling. This work focuses on the voxel pooling process in frustum-based methods, which has rarely been explored but is critical.

\noindent \textbf{Roadside BEV Perception.} Roadside BEV perception is an emerging field, which has been under-explored. BEVHeight \cite{BEVHeight, BEVHeight2} first concentrates on roadside perception, which predicts the height distribution to replace the depth distribution. CBR \cite{CBR} focuses on device robustness, which generates BEV features without extrinsic calibration, while accuracy is limited. CoBEV \cite{CoBEV} fuses geometry-centric depth and semantic-centric height cues to further improve performance. MonoGAE \cite{MonoGAE} considers the prior knowledge of the ground plane. 
Different from these methods, this paper proposes a plug-in to improve the performance of existing frustum-based BEV methods.

\noindent \textbf{Voxel Pooling Strategy.} LSS \cite{LSS} is the pioneering work of frustum-based BEV methods, where voxel pooling is proposed for the first time. 
Extensive work follows this setting \cite{CaDDN, BEVDepth, Bevstereo, BEVHeight}. SA-BEV \cite{SA-BEV} proposes a novel voxel pooling strategy, SA-BEVPool, which filters out background information. While the unfiltered out frustum points adopt the same voxel pooling method as LSS. In this work, we focus on eliminating the position approximation error in the voxel pooling process of LSS.

% \noindent \textbf{Voxel Pooling Strategy.} LSS \cite{LSS} is the pioneering work of frustum-based BEV methods, where voxel pooling is proposed for the first time. 
% Extensive work follows this setting, such as CaDDN \cite{CaDDN}, BEVDepth \cite{BEVDepth}, BEVStereo \cite{Bevstereo} and BEVHeight \cite{BEVHeight}. SA-BEV \cite{SA-BEV} proposes a novel voxel pooling strategy, SA-BEVPool, which filters out background information. While the unfiltered out frustum points adopt the same voxel pooling method as LSS. In this work, we focus on eliminating the position approximation error in the voxel pooling process of LSS.

\section{Methods}
\label{sec:Methods}

In this section, we first give a brief problem formulation of vision-based roadside 3D object detection. Next, an overall architecture of BEVSpread network is presented. Finally, the core designs of BEVSpread are described in detail.

\subsection{Problem Formulation}
In this work, we aim to detect 3D bounding boxes of traffic objects from roadside monocular images. Formally, a 3D object detector can be defined as: 
\vspace{-0.06cm}
\begin{equation}
    B = M_{\theta}(I, E, K)
\vspace{-0.06cm}
\end{equation}
where $M_{\theta}$ is the detection model with the learnable parameters $\theta$, $I \in \mathbb{R}^{H \times W \times 3}$ is the input monocular image, $(H, W)$ represent the height and width of the image, $E \in \mathbb{R}^{3 \times 4}$ and $K \in \mathbb{R}^{3 \times 3}$ are the extrinsic and intrinsic matrix of the roadside camera, respectively. We denote the set of predicted 3D bounding boxes as:
\begin{equation}
    B = \{\hat{B}_1, \hat{B}_2, ..., \hat{B}_n\}
\end{equation}
where $n$ is the number of predicted objects and $\hat{B}$ can be formulated as a vector with 7 degrees of freedom:
\begin{equation}
    \hat{B} = (x, y, z, l, w, h, r)
\end{equation}
where $(x, y, z)$ is the location of the 3D bounding box, $(l, w, h)$ is the length, width and height of the 3D bounding box, and $r$ is the yaw angle relative to one specific axis.

\subsection{BEVSpread}
% \mathcal{C}
\textbf{Overall Architecture.} As shown in \cref{Pipeline}, the overall framework consists of four main stages. The image encoder is composed of a ResNet \cite{Resnet} and a SECONDFPN \cite{Second}, aiming to extract the 2D high-dimensional multi-scale image features $f^{\mathcal{I}} \in \mathbb{R}^{C_\mathcal{I} \times \frac{H}{16} \times \frac{W}{16}}$ from a monocular roadside image $I$, where $C_\mathcal{I}$ denotes the channel number. The depth/height projector first takes the 2D image features $f^{\mathcal{I}}$ and camera parameters ($E$, $K$) as input to predict the depth/height distribution $f^{\mathcal{D}} \in \mathbb{R}^{C_\mathcal{D} \times \frac{H}{16} \times \frac{W}{16}}$ and the context features $f^{\mathcal{C}} \in \mathbb{R}^{C_\mathcal{C} \times \frac{H}{16} \times \frac{W}{16}}$, where $C_\mathcal{D}$ represents the number of depth/height bins and $C_\mathcal{C}$ stands for the channels of the context features. These two are further fused through an outer product operation to obtain 2.5D frustum features $f^{2.5D} \in \mathbb{R}^{C_\mathcal{C} \times C_\mathcal{D} \times \frac{H}{16} \times \frac{W}{16}}$. Then, the projector push the 2.5D frustum features $f^{2.5D}$ into 3D geometry features $f^{3D} \in \mathbb{R}^{X \times Y \times Z \times C_\mathcal{C}}$ using camera parameters ($E$, $K$). The proposed spread voxel pooling strategy splattes the 3D geometry features $f^{3D}$ into an unified BEV features $f^{\text{BEV}}$. Finally, the 3D detection head utilizes the generated BEV features to produce the 3D bounding boxes $B$. 

\vspace{0.5em} \noindent \textbf{Top-k Nearest BEV Grids.} Define $P^{\text{BEV}}$ to represent the set of arbitrary positions in BEV grids, $\dot{P}^{\text{BEV}} \in P^{\text{BEV}}$ to represent the set of BEV grid centers, $\Omega_{p,k} \subseteq \dot{P}^{\text{BEV}}$ to represent the set of top-$k$ neareast BEV grid centers to $p=(x,y) \in P^{\text{BEV}}$. For $\forall \hat{p}=(\hat{x},\hat{y}) \in \Omega_{p,k}$ and $\Bar{p}=(\Bar{x}, \Bar{y}) \in \{\dot{P}^{\text{BEV}} \backslash \Omega_{p,k}\}$, it should satisfies:
\begin{equation}
\left\{
             \begin{array}{lr}
             |\Omega_{p,k}|=k, &  \\
             % y=s, & 0\leq s\leq L,|t|\leq1.\\
             d_{p, \hat{p}} \leq d_{p, \Bar{p}}&  
             \end{array}
\right.
\label{equation:Omega}
\end{equation}
\begin{equation}
    d_{p, p'} = \sqrt{(x-x')^2 + (y-y')^2}
    \label{distance}
\end{equation}
where $|\cdot|$ denotes the cardinality of a set, $k$ represents the neighbors number of $\forall p \in P^{\text{BEV}}$, $\{\dot{P}^{\text{BEV}} \backslash \Omega_{p,k}\}$ denotes the relative complement of $\Omega_{p,k}$ in $\dot{P}^{\text{BEV}}$, and $d_{p, p'}$ represents the Euclidean distance between $p = (x,y) \in P^{\text{BEV}}$ and $p'=(x', y') \in P^{\text{BEV}}$.

\vspace{0.5em} \noindent \textbf{Spread Voxel Pooling.} In the spread voxel pooling stage, we first calculate the corresponding positions $p = (x, y) \in P^\text{BEV}$ in BEV space for each point $(x, y, z) \in P^{3D}$ in 3D geometry by reducing the $Z$-axis degree of freedom. Instead of accumulating the included context feature $f^\mathcal{C} \in \mathbb{R}^{C_\mathcal{C}}$ of $p$ into the corresponding single BEV gird center, we propagate $f^\mathcal{C}$ with certain weights to its neighbors $\Omega $, which are the $n$ nearest BEV grids center around $p$. % The neighbor selection process is illustrated in \cref{neighbor}. 
Specifically, the process of spread voxel pooling can be formulated as:
\begin{equation}
    f^\text{BEV}_{\hat{p}} = \text{Add}(f^\text{BEV}_{\hat{p}}, \omega_{p, \hat{p}}  \cdot f^\mathcal{C}_p ), \forall \hat{p} \in \Omega_{p, k}
    \label{BEV_feat}
\end{equation}
where $\Omega_{p,k}$ is the set of top-$k$ neareast BEV grid centers of $p$, $f^\text{BEV}_{\hat{p}}$ denotes the BEV feature of $\hat{p}$, $\omega_{p, \hat{p}}$ represents the weight of $\hat{p}$ determined by the weight decay function, $f^\mathcal{C}$ is the context feature included in $p$, and $\text{Add}(a,b) = a + b$.

\begin{figure}[!t]
    \centering
    \includegraphics[width=1.0\linewidth]{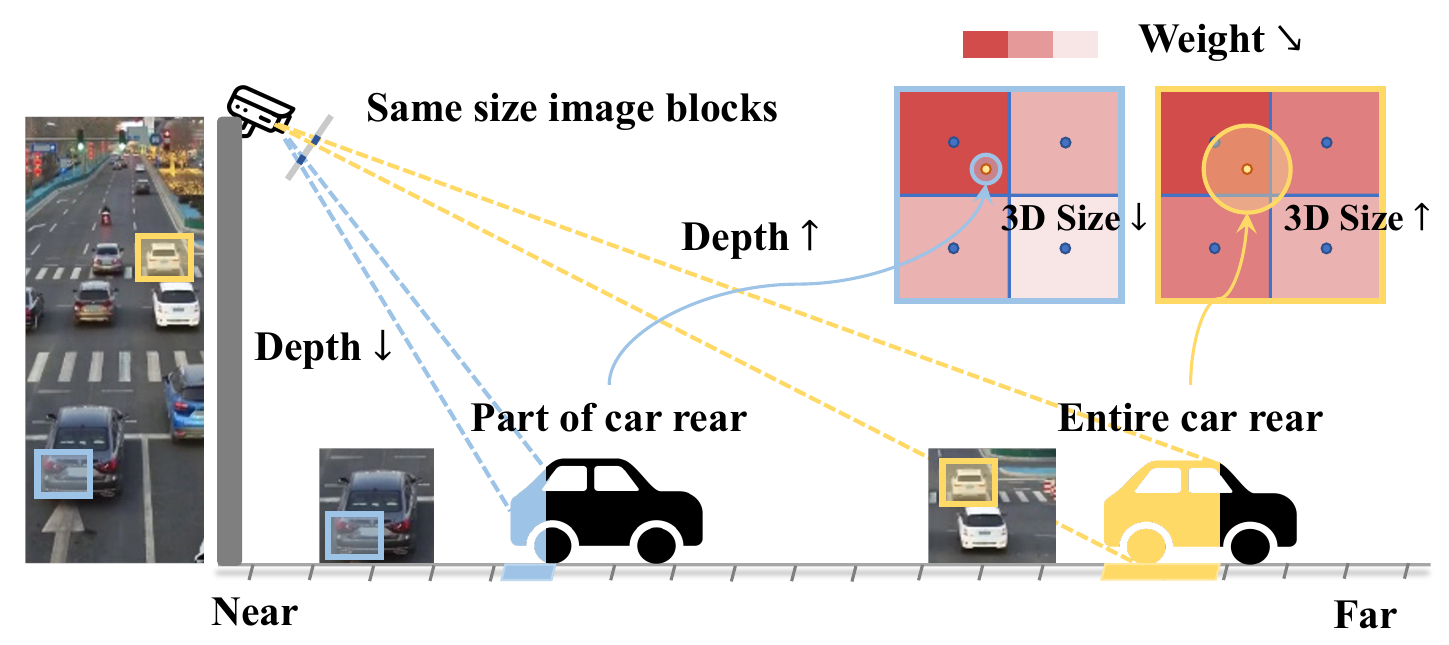}
    \vspace{-0.7cm}
    \caption{\textbf{Effect of depth in voxel pooling.} Same size image blocks with deeper depth represent objects of larger 3D scales, which results in distant objects containing few image features. Therefore, it is reasonable to assign larger weights to the surrounding BEV grids for the distant targets.}
    \vspace{-0.3cm}
    \label{depth_effect}
\end{figure}

\vspace{0.5em} \noindent \textbf{Weight Function.} We discover that the weights should be related to the distance and depth in spread process. \textbf{(a)} Weight decay with distance can retain more location information, which is beneficial to recover the accurate position of $p \in P^\text{BEV}$ through subsequent network learning, so as to eliminate the position approximation error in the original voxel pooling process. Additionally, we have designed an intuitive experiment to demonstrate this point in \cref{prove_exper}. \textbf{(b)} As shown in \cref{depth_effect}, same size image blocks with deeper depth represent objects of larger 3D scales, resulting in distant objects containing few image features. Therefore, it is reasonable to assign larger weights to the surrounding BEV grids for distant targets, manifesting that the weights decay more slowly with distance, as shown in \cref{Pipeline}. 

To this end, we design a specific weight function, which ingeniously utilizes a Gaussian function to integrate the distance and depth information. The function is defined as:
% spreads features with adaptive weights according to depth and distance. The function is defined as:
\begin{equation}
    \omega_{p, \hat{p}} = \text{exp}({\frac{-{d^2_{p, \hat{p}}}}{\sigma^2}})
    \label{Fw}
\end{equation}
\begin{equation}
    \sigma^2 = \alpha \cdot D_p 
\end{equation}
where $\omega_{p, \hat{p}}$ represents the calculated weight of $\hat{p}$, $d_{p, \hat{p}}$ represents the the Euclidean distance between $p$ and $\hat{p}$, $D_p$ is the predicted depth of $p$, $\sigma^2$ is the variance of Gaussian function which is positively related to $D_p$ and controls the decay speed of $\omega_{p, \hat{p}}$, and $\alpha$ is a learnable parameter to maintain $\sigma^2$ within the interval [0,2]. 
Through this function, the weights change adaptively depending on the distance and depth. 
% Compared with Gaussian function, our function achieves superior performance. 
In summary, the pseudocode of the spread voxel pooling strategy is shown in \cref{algorithm}.

\begin{algorithm}[!t]
\small
\caption{Spread Voxel Pooling}
\label{algorithm}
\textbf{INPUT}: 3D geometry points $P^{3D} \in \mathbb{R}^{X \times Y \times Z}$, context image feature of each 3D geometry point $f^\mathcal{C} \in \mathbb{R}^{C_\mathcal{C}}$, depth vector of 3D geometry points $D$.\\
% $\mathcal{L} \times \mathcal{W}$: BEV grids size.\\
\textbf{OUTPUT}: BEV features $f^\text{BEV}$.\\
\textbf{BEGIN}: 
\begin{algorithmic}[1] 
% \STATE $\triangleright$ \textit{\textbf{Initialization}} 
% \vspace{1mm}
% \STATE $P^{3D} \in \mathbb{R}^{X \times Y \times Z}$ and $f^\mathcal{C} \in \mathbb{R}^{C_\mathcal{C}}$ extracted from $f^{3D}$
\STATE $P^\text{BEV} \in \mathbb{R}^{X \times Y}$ extracted from $P^{3D}$
% \STATE $f^\text{BEV} = 0^{\mathcal{L} \times \mathcal{W} \times C_\mathcal{C}}$
% \vspace{1mm}
\FOR{$p$ in $P^\text{BEV}$}
% \STATE \textcolor{magenta}{$\triangleright$} \textit{\textcolor{magenta}{Select Top-$k$ Neareast BEV Grids, $k>1$}}
% \STATE Select neighbors $\Omega_{p, k}$ (Spread) \textcolor[RGB]{34,139,34}{$\leftrightarrow$ $\Omega_{p, 1}$ (LSS)}
\STATE Get $\Omega_{p, k}$ by Eq.~(\ref{equation:Omega}) \hfill \textit{\textcolor{magenta}{$\triangleright$ Top-$k$ Neareast BEV Grids}}
% \STATE Get $\Omega_{p, k}$ by Eq.~(\ref{equation:Omega}) \hfill \textit{\textcolor[RGB]{34,139,34}{$\triangleright$ Top-$k$ Neareast BEV Grids}}
\FOR{$\hat{p}$ in $\Omega_{p, k}$ }
\STATE $\omega_{p, \hat{p}} \leftarrow \text{exp}({\frac{-{d_{p, \hat{p}}}}{\alpha \cdot D_p}})$ \hfill \textcolor{magenta}{$\triangleright$} \textit{\textcolor{magenta}{Weight Calculation}}
% \textcolor[RGB]{34,139,34}{$\leftrightarrow$ $\omega_{p, \hat{p}} = 1.0$ (LSS)}
\STATE $f^\text{BEV}_{\hat{p}} \leftarrow \text{Add}(f^\text{BEV}_{\hat{p}}, \omega_{p, \hat{p}}  \cdot f^\mathcal{C}_p )$ \hfill \textcolor{magenta}{$\triangleright$} \textit{\textcolor{magenta}{Feature Accumulation}}
\ENDFOR
\ENDFOR
\RETURN $f^\text{BEV}$
\end{algorithmic}
\textbf{END}
\end{algorithm}
\section{Experiments}
\label{sec:Experiments}

In this section, we first introduce two roadside benchmark datasets and the implementation details. Then, we compare our proposed BEVSpread with state-of-the-art methods. Finally, comprehensive ablation studies are conducted to validate the effects of each component. 

\subsection{Datasets}

\noindent \textbf{DAIR-V2X-I.} DAIR-V2X \cite{DAIR-V2X} is a large-scale dataset for vehicle-infrastructure cooperative autonomous driving, which offers a multi-modal 3D object detection resource. Here, we focus on DAIR-V2X-I subset, containing $10k$ images from mounted cameras to study roadside perception. DAIR-V2X-I involves $493k$ 3D bounding box annotations, spanning distances from 0 to 200 meters. Following the previous work \cite{BEVHeight}, 50\%, 20\% and 30\% images are split into train, validation, and testing, respectively. Noting that the testing set is not yet published and we evaluate the results on the validation set.

\vspace{0.5em} \noindent \textbf{Rope3D.} Rope3D \cite{Rope3D} is another benchmark for roadside 3D object detection, consisting of $50k$ images and over $1.5M$ 3D objects collected across a variety of lighting conditions (daytime / night / dusk), different weather conditions (rainy / sunny / cloudy) and 26 distinct intersections, spanning distances from 0 to 200 meters. Following the split strategy detailed in Rope3D, we use 70\% of the images as training, and the remaining 30\% as testing. % Notably, LiDAR data is not publicly accessible in Rope3D.

\vspace{0.5em} \noindent \textbf{Metrics.} For both DAIR-V2X-I and Rope3D datasets, we report the 40-point average precision ($\text{AP}_{\text{3D}|\text{R40}}$) \cite{AP3D} of 3D bounding boxes, which is further categorized into three modes: Easy, Middle and Hard, based on the box characteristics, including size, occlusion and truncation, following the metrics of KITTI \cite{KITTI}. 
% Additionally, we focus on three classes of targets: Vehicle, Pedestrian and Cyclist.

\subsection{Implementation Details}
For fair comparison with state-of-the-art methods, we use ResNet-101 \cite{Resnet} as image encoder, BEV grid size is set to 0.4 meters, the range of X axis is set to 0-100 meters, and the neighbors number is set to 6. ResNet-50 and 0.8m grid size are used for ablation studies. Following BEVHeight \cite{BEVHeight}, we adopt image data augmentations including random intrinsic and extrinsic changes. We use AdamW \cite{AdamW} as an optimzer with a learning rate set to $2e-4$. All experiments are conducted on 8 RTX-3090 GPUs.

\subsection{Comparison with state-of-the-art}
For a comprehensive evaluation, we compare the proposed BEVSpread with state-of-the-art BEV detectors on DAIR-V2X-I and Rope3D. Since the proposed spread voxel pooling strategy is a plug-in, we deploy it to BEVHeight, dubbed BEVSpread. The results are described as follows.

\vspace{0.5em} \noindent \textbf{Results on DAIR-V2X-I.} \cref{tab:dair} illustrates the performance comparison on DAIR-V2X-I. We compare our BEVSpread with the state-of-the-art vision-based methods, including ImVoxelNet \cite{ImVoxelNet}, BEVFormer \cite{BEVFormer}, BEVDepth \cite{BEVDepth} and BEVHeight \cite{BEVHeight}, and the traditional LiDAR-based methods, including PointPillars \cite{PointPillars}, SECOND \cite{Second} and MVXNet \cite{MVXNet}. The results demonstrate that BEVSpread outperforms state-of-the-art methods by a significant margin of (1.12, 5.26 and 3.01) AP in vehicle, pedestrian, and cyclist categories, respectively. We notice that previous methods are trained only in 0-100m, while DAIR-V2X-I contains the labels of 0-200m. To this end, we cover a longer range of 3D object detection, locating targets in 0-200m, which is denoted as DAIR-V2X-I$^*$ in \cref{tab:dair}.

\begin{table*}[!h]
    \renewcommand{\arraystretch}{1.1}
    \centering
    \caption{\textbf{Comparison $\text{AP}_{\text{3D}|\text{R40}}$ results of 3D object detection on the validation set of DAIR-V2X-I~\cite{DAIR-V2X} and Rope3D~\cite{Rope3D} }. ResNet-101 is used as image encoder, the BEV grid size is set to 0.4 meters, and $\text{top-}k$ ($k$=6) nearest BEV grid centers are selected as neighbors. ``$*$" denotes covering the longer range between $0 {\sim} 200m$, while others cover $0 {\sim} 100m$.
    }
    \label{tab:dair}
    \resizebox{1.0\textwidth}{!}{

    \begin{tabular}{llcc|ccc|ccc|ccc}
        \toprule
        \multirow{2}{*}{\textbf{Dataset}} & \multirow{2}{*}{\textbf{Method}} & \multirow{2}{*}{\textbf{Modality}} & \multirow{2}{*}{\textbf{Venue}} & \multicolumn{3}{c|}{\textbf{Vehicle (\textit{IoU=0.5})}} & \multicolumn{3}{c|}{\textbf{Pedestrian (\textit{IoU=0.25})}} & \multicolumn{3}{c}{\textbf{Cyclist (\textit{IoU=0.25})}} \\
        & & & & Easy & Middle & Hard & Easy & Middle & Hard & Easy & Middle & Hard \\
        
        \Xhline{0.75pt}
        
        \rowcolor{gray!15} DAIR-V2X-I~\cite{DAIR-V2X} & & & & & & & & & & & & \\
        & PointPillars~\cite{PointPillars} & LiDAR & CVPR' 19 &63.07&54.00&54.01&38.53&37.20&37.28&38.46&22.60&22.49\\
        & SECOND~\cite{Second} & LiDAR & Sensors &71.47&53.99&54.00&55.16&52.49&52.52&54.68&31.05&31.19\\
        & MVXNet~\cite{MVXNet} & LiDAR \& Camera & ICRA' 19 &71.04&53.71&53.76&55.83&54.45&54.40&54.05&30.79&31.06\\
        & ImVoxelNet~\cite{ImVoxelNet} & Camera & WACV' 22 &44.78&37.58&37.55&6.81&6.75&6.74&21.06&13.57&13.17\\
        & BEVFormer~\cite{BEVFormer} & Camera & ECCV' 22 &61.37&50.73&50.73&16.89&15.82&15.95&22.16&22.13&22.06\\
        & BEVDepth~\cite{BEVDepth} & Camera & AAAI' 23 &75.50&63.58&63.67&34.95&33.42&33.27&55.67&55.47&55.34\\
        & BEVHeight~\cite{BEVHeight} & Camera & CVPR' 23 &77.78&65.77&65.85&41.22&39.29&39.46&60.23&60.08&60.54\\
        & BEVSpread (\textbf{Ours})& Camera & - & \textbf{79.07} & \textbf{66.82} & \textbf{66.88} & \textbf{46.54} & \textbf{44.51} & \textbf{44.71} & \textbf{62.64} & \textbf{63.50} & \textbf{63.75}\\ 
        & \textit{w.r.t. BEVHeight} &  &  & \textcolor{red}{+1.29} & \textcolor{red}{+1.05} & \textcolor{red}{+1.03} & \textcolor{red}{+5.32}& \textcolor{red}{+5.22}& \textcolor{red}{+5.25}& \textcolor{red}{+2.41}& \textcolor{red}{+3.42} & \textcolor{red}{+3.21}\\
        
        \Xhline{0.75pt}
        
        \rowcolor{gray!15} DAIR-V2X-I$^*$~\cite{DAIR-V2X} & & & & & & & & & & & & \\ 
         & BEVHeight~\cite{BEVHeight} & Camera & CVPR 23 & 81.62 & 75.90 & 75.94  & 40.89 & 38.98 & 39.18 & 60.29 & 60.60 & 61.13 \\
         & BEVSpread (\textbf{Ours})& Camera & - & \textbf{82.84} & \textbf{77.10} & \textbf{77.19} & \textbf{43.96} & \textbf{42.03} & \textbf{42.13} & \textbf{62.31} & \textbf{64.44} & \textbf{64.89}\\
        & \textit{w.r.t. BEVHeight} & & & \textcolor{red}{+1.22}&\textcolor{red}{+1.21}&\textcolor{red}{+1.25}&\textcolor{red}{+3.07}&\textcolor{red}{+3.05}&\textcolor{red}{+2.95}&\textcolor{red}{+2.02}&\textcolor{red}{+3.84}&\textcolor{red}{+3.76}\\
        
        \Xhline{0.75pt}
        
        \rowcolor{gray!15} Rope3D~\cite{Rope3D} & & & & & & & & & & & & \\
         & BEVDepth~\cite{BEVDepth} & Camera & AAAI 23 & 76.90 & 66.91 & 66.89 & 30.42 & 28.08 & 28.11 & 55.34 & 53.53 & 53.51\\
         & BEVHeight~\cite{BEVHeight} & Camera & CVPR 23 & 77.93 & 67.50 & 67.49 & 36.26 & 30.35 & 30.30 & 61.49 & 56.98 & 56.90\\
         & BEVSpread (\textbf{Ours})& Camera & - & \textbf{80.61} & \textbf{70.04} & \textbf{70.03} & \textbf{38.65} & \textbf{34.32} & \textbf{34.25} & \textbf{63.66} & \textbf{59.11} & \textbf{59.03}\\
         & \textit{w.r.t. BEVHeight} & & & \textcolor{red}{+2.69}&\textcolor{red}{+2.55}&\textcolor{red}{+2.54}&\textcolor{red}{+2.39}&\textcolor{red}{+3.97}&\textcolor{red}{+3.95}&\textcolor{red}{+2.17}&\textcolor{red}{+2.13}&\textcolor{red}{+2.13}\\
        \bottomrule
    \end{tabular}

    }
\end{table*}
\begin{figure*}[!h]
    \centering
    \includegraphics[width=1\linewidth]{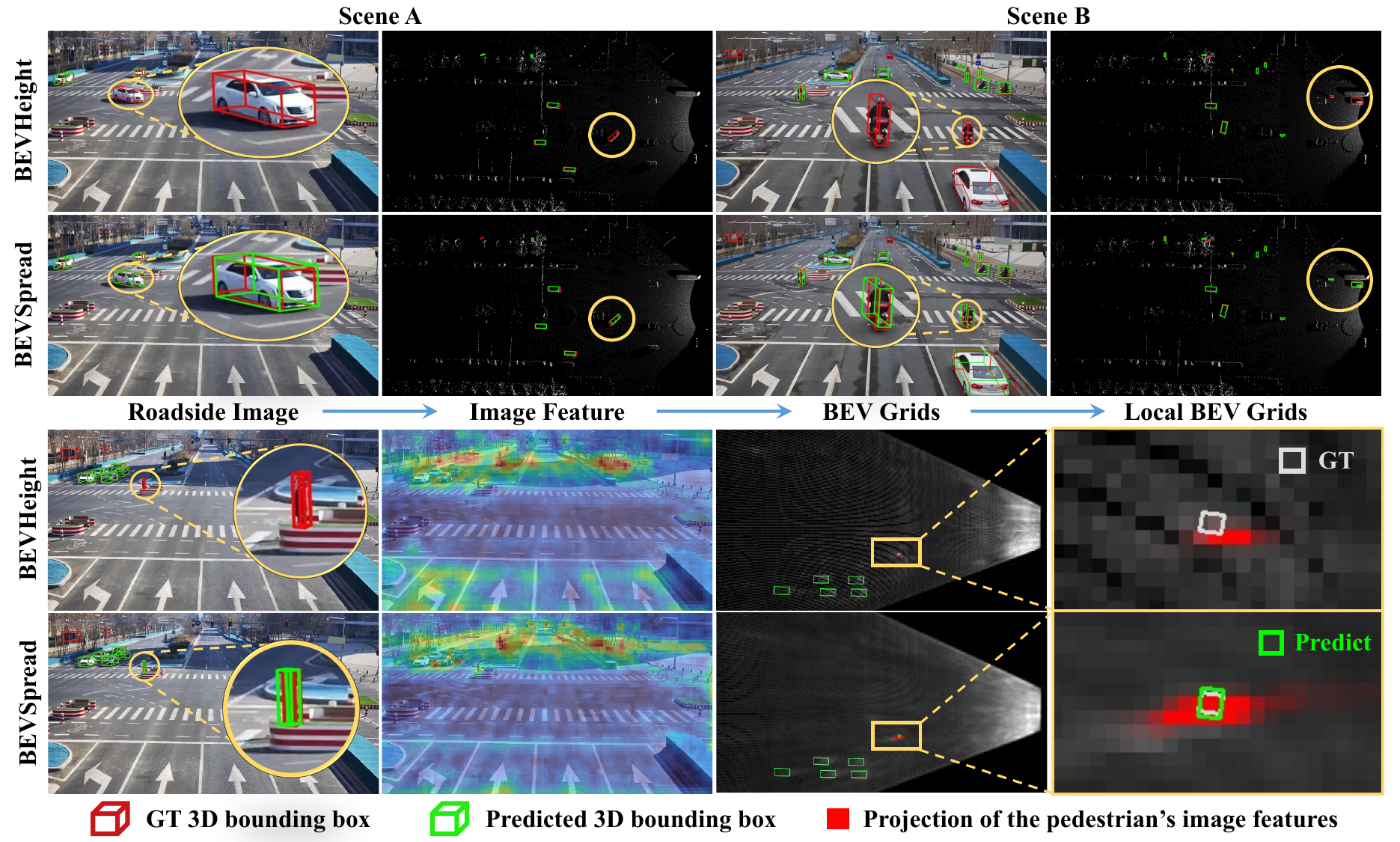}
    \caption{\textbf{Visualization results of BEVHeight and our proposed BEVSpread in image and BEV view.} It can be observed in the upper half that BEVSpread detects the targets which BEVHeight have not detected in multiple scenes. The lower half demonstrates the reasons. We notice that BEVHeight misses the pedestrian because no corresponding image features are projected onto the correct BEV grids. However, BEVSpread spreads the image features to the surrounding BEV grids and thus successfully detects the target.}
    \label{Visualization}
\end{figure*}

\vspace{0.5em} \noindent \textbf{Results on Rope3D.} We compare our BEVSpread with the state-of-the-art vision-centric methods, including BEVDepth \cite{BEVDepth} and BEVHeight \cite{BEVHeight}, on the Rope3D validation set in homologous settings. As shown in \cref{tab:dair}, BEVSpread outperforms all other methods across the board, with significant improvements of (2.59, 3.44 and 2.14) AP in vehicle, pedestrian, and cyclist, respectively.

\vspace{0.5em} \noindent \textbf{Visualization Results.} As shown in \cref{Visualization}, we present the visualization results of BEVHeight \cite{BEVHeight} and our BEVSpread in the image and BEV view. It can be observed in the upper half that BEVSpread detects the targets which BEVHeight misses in multiple scenes. The main reason is shown in the lower half. Image features show that BEVSpread focuses more attention on the foreground area. And the BEV features generated by BEVSpread are smoother than those generated by BEVHeight. BEVHeight misses the pedestrian because no corresponding image features are projected onto the correct BEV grids. While BEVSpread spreads the image features to the surrounding BEV grids and exactly covers the correct BEV grids, so as to successfully detect the target.

\subsection{Results on nuScenes.}
\label{nuscenes_results}
Our approach specifically targets the roadside scenario. To further assess its robustness, we conduct additional experiment on nuScenes following BEVDepth \cite{BEVDepth}. \cref{tab:nuScenes} shows that BEVSpread still works in ego-vehicle settings, and the improvement (4.2\% NDS) is comparable to that in roadside scenario (5.5\% Avg-AP).

\subsection{Proof Experiment for Position Recovery}
\label{prove_exper}
\begin{figure}[h]
    \centering
    \includegraphics[width=0.9\linewidth]{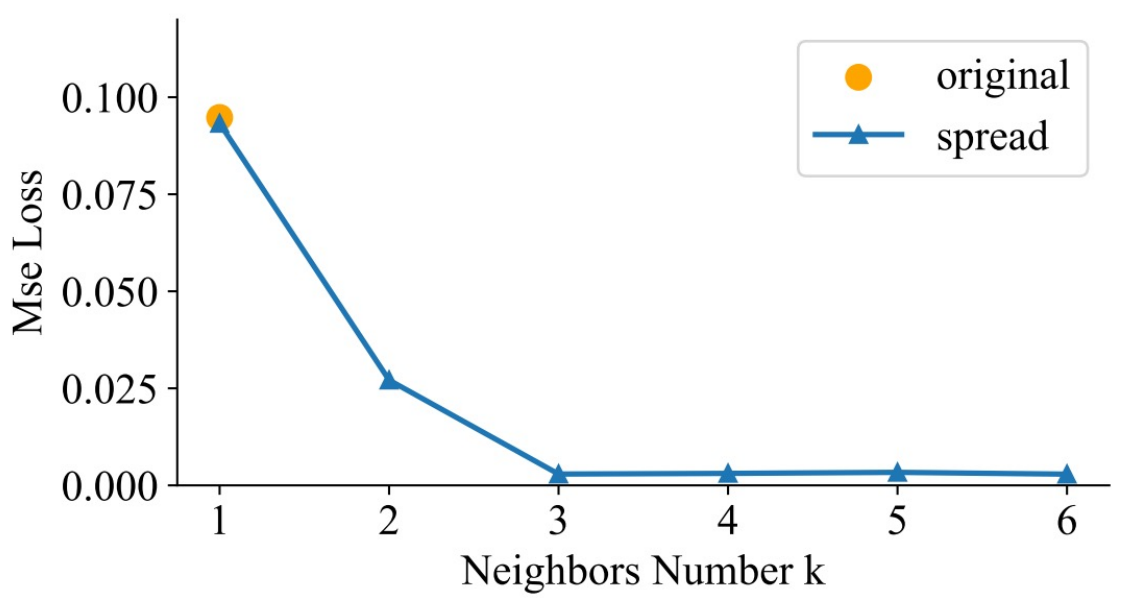}
    \vskip -1ex
    \caption{\textbf{Proof Experiment for Position Recovery.} Spread voxel pooling recovers the random point position with 0.003 MSE loss when the neighbors number $k \geq 3$, while the original voxel pooling ($k=1$) obtains 0.095 MSE loss.}
    \label{prove}
    \vskip -3ex
\end{figure}

We have designed an intuitive experiment to demonstrate that the proposed spread voxel pooling strategy can achieve accurate position recovery in BEV space. Initially, 10 random vectors of $C$ dimensions representing image features are randomly generated. Then, we randomly generate 3D points and assign for these 10 features. Based on the original voxel pooling and spread voxel pooling, the 3D points are projected onto the $16 \times 16$ BEV grids to obtain the BEV features. The U-Net encoder network is utilized to regress the accurate position of the first image feature in the BEV space, and MSE loss is used. Note that the training process contains 5,000 iterations, and the batch size is set to 128 per iteration. The inputs are random for each iteration. The experimental process is shown in \cref{first_img}. As shown in \cref{prove}, our spread voxel pooling recovers the random point position with 0.003 MSE loss when the neighbors number $\geq 3$, while the original voxel pooling obtains 0.095 MSE loss. 

\begin{table}[!t]
\renewcommand{\arraystretch}{1.2}
\huge
  \centering
    \resizebox{0.48 \textwidth}{!}{
  \begin{tabular}{l|ccccccc}
    \toprule
    \textbf{Method} & \textbf{NDS $\uparrow$} & \textbf{mAP $\uparrow$}  & \textbf{mATE $\downarrow$} & \textbf{mASE $\downarrow$} & \textbf{mAOE $\downarrow$} & \textbf{mAVE $\downarrow$} & \textbf{mAAE $\downarrow$} \\
    \midrule
     \textcolor[RGB]{180,180,180}{BEVDepth}   & \textcolor[RGB]{180,180,180}{0.436} & \textcolor[RGB]{180,180,180}{0.330} & \textcolor[RGB]{180,180,180}{0.702} & \textcolor[RGB]{180,180,180}{0.280} & \textcolor[RGB]{180,180,180}{0.535} & \textcolor[RGB]{180,180,180}{0.553} & \textcolor[RGB]{180,180,180}{0.227}  \\
     \midrule
     BEVDepth$^*$              & 0.432 & 0.325 & 0.701 & 0.283 & 0.572 & 0.531 & 0.224   \\
     BEVDepth$^*$ + ours       & \textbf{0.450} & \textbf{0.327} & \textbf{0.688} & \textbf{0.275} & \textbf{0.489} & \textbf{0.470} & \textbf{0.217}   \\     
    \bottomrule
  \end{tabular}
  }
  \vspace{-.2cm}
  \caption{Comparison on the nuScenes $val$ set. The experiment is reproduced based on the official BEVDepth repository with config named bev\_depth\_lss\_r50\_256x704\_128x128\_24e\_2key. 
  Both CBGS and EMA are not used. \textcolor[RGB]{180,180,180}{BEVDepth} denotes the official result of this config. $^*$ denotes the results we reproduce.}
  \vspace{-.4cm}
  \label{tab:nuScenes}
\end{table}

\begin{figure}[h]
    \centering
    \includegraphics[width=0.9\linewidth]{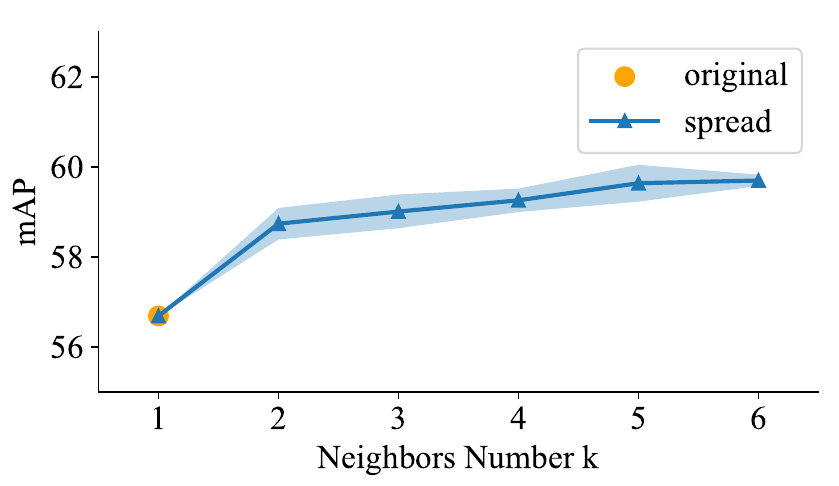}
    \vskip -1ex
    \caption{\textbf{Hyperparameter sensitivity experiment on neighbors number $k$.} It can be observed that the performance of $k \geq 2$ is significantly better than $k=1$ (baseline). As $k$ increases, the performance gradually improves and becomes stable.}
    \label{neighbor}
    \vskip -2ex
\end{figure}

\subsection{Ablation Study}
\label{ablation}
\vspace{0.5em} \noindent \textbf{Performance as a plugin.} The proposed spread voxel pooling strategy, as a plug-in, can significantly improve the performance of existing frustum-based BEV methods. As shown in \cref{tab:dair_spread}, after being deployed to BEVDepth \cite{BEVDepth}, the performance has been significantly improved by a margin of (4.17, 8.93 and 8.2) AP in three categories. After being deployed to BEVHeight \cite{BEVHeight}, the performance has been improved by a margin of (1.55, 5.58 and 7.56) AP in three categories. It is worth noting that the recognition ability for pedistrian and cyclist has been greatly improved.

\vspace{0.5em} \noindent \textbf{Analysis on Neighbor Selection.} 
\cref{neighbor} shows how the mAP of three categories changes with neighbors number $k$. For each hyperparameter selection, we repeat 3 times, and the light-blue area indicates the error range. It can be observed that the performance of $k \geq 2$ is significantly better than $k=1$ (baseline). As $k$ increases, performance gradually improves and becomes stable. 

\vspace{0.5em} \noindent \textbf{Analysis on Weight Function.}
We validate the effectiveness of the depth and the learnable parameter $\alpha$ for weight function in \cref{tab:ablation_study}. The improvement in three major categories proves that the application of depth and the learnable parameter $\alpha$ allows for better spread performance. When applying both, considerable performance (65.80\%, 31.00\%, 56.34\%) is gained for three categories in middle difficulty.

\begin{table*}[!t]
    \setlength{\tabcolsep}{4mm} % column spacing
    % \small
    \renewcommand{\arraystretch}{1.1}
    \centering
    \caption{\textbf{Ablation study of spread voxel pooling on the DAIR-V2X-I \cite{DAIR-V2X}.} ResNet-50 is used as image encoder, the BEV grid size is set to 0.8 meters, and the detection range is set to $0 {\sim} 100m$, and $\text{top-}k$ ($k$=2)  nearest BEV grid centers are selected as neighbors.}
    \label{tab:dair_spread}
    \vskip -1ex
    \resizebox{1.0\textwidth}{!}{

    \begin{tabular}{l|ccc|ccc|ccc}
        \toprule
        
        \multirow{2}{*}{\textbf{Method}} &  \multicolumn{3}{c|}{\textbf{Vehicle (\textit{IoU=0.5})}} & \multicolumn{3}{c|}{\textbf{Pedestrian (\textit{IoU=0.25})}} & \multicolumn{3}{c}{\textbf{Cyclist (\textit{IoU=0.25})}} \\
        & Easy & Middle & Hard & Easy & Middle & Hard & Easy & Middle & Hard \\
        
        \Xhline{0.75pt}
        
        BEVDepth~\cite{BEVDepth} 
        &71.09&60.37&60.46&21.23&20.84&20.85&40.54&40.34&40.32\\
        % \rowcolor{gray!15}
        + spread voxel pooling 
        &\textbf{76.15}&\textbf{64.09}&\textbf{64.19}&\textbf{30.87}&\textbf{29.27}&\textbf{29.57}&\textbf{48.06}&\textbf{48.53}&\textbf{49.21}\\
        % \rowcolor{gray!15}
        \textit{w.r.t. BEVDepth}
        &\textcolor{red}{+5.06}&\textcolor{red}{+3.72}&\textcolor{red}{+3.73}&\textcolor{red}{+9.64}&\textcolor{red}{+8.43}&\textcolor{red}{+8.72}&\textcolor{red}{+7.52}&\textcolor{red}{+8.19}&\textcolor{red}{+8.89}\\
        
        \hline
        
        BEVHeight~\cite{BEVHeight}
        &76.24&64.54&64.13&26.47&25.79&25.72&48.55&48.21&47.96\\
        % \rowcolor{gray!15}
        + spread voxel pooling 
        & \textbf{77.91}  & \textbf{65.80} &\textbf{65.86} & \textbf{32.48} & \textbf{31.00} & \textbf{31.25} & \textbf{54.19} & \textbf{56.34} & \textbf{56.88} \\
        % \rowcolor{gray!15}
        \textit{w.r.t. BEVHeight}
        &\textcolor{red}{+1.67}&\textcolor{red}{+1.26}&\textcolor{red}{+1.73}&\textcolor{red}{+6.01}&\textcolor{red}{+5.21}&\textcolor{red}{+5.53}&\textcolor{red}{+5.64}&\textcolor{red}{+8.13}&\textcolor{red}{+8.92}\\
        
        \bottomrule
    \end{tabular}

    }
    \vskip -0.3ex
\end{table*}
\begin{table*}[!t]
    % \small
    \setlength{\tabcolsep}{1.3mm} % column spacing
    \renewcommand{\arraystretch}{1.4}
    \centering
    \caption{\textbf{Ablation study of weight function on DAIR-V2X-I \cite{DAIR-V2X}.} ResNet-50 is used as image encoder, the BEV grid size is set to 0.8 meters, and the detection range is set to $0 {\sim} 100m$, and $\text{top-}k$ ($k$=2) nearest BEV grid centers are selected as neighbors.} %The value in brackets denotes the discrepancy to baseline.}
    \label{tab:ablation_study}
    \vskip -1ex
    \resizebox{1.0\textwidth}{!}{
        
    \begin{tabular}{ccc|ccc|ccc|ccc}
        \toprule
        
        \multicolumn{3}{c|}{\textbf{Spacing}}  & \multicolumn{3}{c|}{\textbf{Vehicle (\textit{IoU=0.5})}} & \multicolumn{3}{c|}{\textbf{Pedestrian (\textit{IoU=0.25})}} & \multicolumn{3}{c}{\textbf{Cyclist (\textit{IoU=0.25})}} \\
        spread & Depth & $\alpha$ & Easy & Middle & Hard & Easy & Middle & Hard & Easy & Middle & Hard\\
        
        \Xhline{0.75pt}
        
        -&-&-&76.24&64.54&64.13&26.47&25.79&25.72&48.55&48.21&47.96\\
        \checkmark&-&-&77.67 \textcolor{red}{(+1.43)}&65.61 \textcolor{red}{(+1.07)}&65.69 \textcolor{red}{(+1.56)}&31.34 \textcolor{red}{(+4.87)}&29.94 \textcolor{red}{(+4.15)}&30.08 \textcolor{red}{(+4.36)}&53.53 \textcolor{red}{(+4.98)}&54.65 \textcolor{red}{(+6.44)}&55.17 \textcolor{red}{(+7.21)}\\
        \checkmark&\checkmark&-&77.88 \textcolor{red}{(+1.64)}&65.79 \textcolor{red}{(+1.25)}&65.76 \textcolor{red}{(+1.63)}&32.40 \textcolor{red}{(+5.93)}&30.97  \textcolor{red}{(+5.18)}&31.18 \textcolor{red}{(+5.46 )}&53.69 \textcolor{red}{(+5.14)}&55.54 \textcolor{red}{(+7.33)}&55.46 \textcolor{red}{(+7.50)}\\
        \checkmark&-&\checkmark&77.71 \textcolor{red}{(+1.47)}&65.66 \textcolor{red}{(+1.12)}&65.74 \textcolor{red}{(+1.61)}&31.72 \textcolor{red}{(+5.25)}&30.31 \textcolor{red}{(+4.52)}&30.52 \textcolor{red}{(+4.80)}&53.97 \textcolor{red}{(+5.42)}&55.64 \textcolor{red}{(+7.43)}&55.58 \textcolor{red}{(+7.62)}\\
        \checkmark&\checkmark&\checkmark &\textbf{77.91}  \textcolor{red}{(+1.67)}&\textbf{65.80}  \textcolor{red}{(+1.26)}&\textbf{65.86}  \textcolor{red}{(+1.73)}&\textbf{32.48}  \textcolor{red}{(+6.01)}&\textbf{31.00}  \textcolor{red}{(+5.21)}&\textbf{31.25}  \textcolor{red}{(+5.53)}&\textbf{54.19}  \textcolor{red}{(+5.64)}&\textbf{56.34}  \textcolor{red}{(+8.13)}&\textbf{56.88}  \textcolor{red}{(+8.92)}\\
        
        \bottomrule
    \end{tabular}

    }
    \vskip -0.3ex
\end{table*}

\begin{table*}[!t]
    \setlength{\tabcolsep}{4mm} % column spacing
    % \small
    \renewcommand{\arraystretch}{1.1}
    \centering
    \caption{\textbf{Ablation study of different backbones on the DAIR-V2X-I \cite{DAIR-V2X}.} ConvNeXt-B is used as image encoder, the BEV grid size is set to 0.4 meters, and the detection range is set to $0 {\sim} 100m$, and $\text{top-}k$ ($k$=4)  nearest BEV grid centers are selected as neighbors.}
    \label{tab:backbones}
    \vskip -1ex
    \resizebox{1.0\textwidth}{!}{
    \begin{tabular}{l|ccc|ccc|ccc}
        \toprule
        
        \multirow{2}{*}{\textbf{Method}} &  \multicolumn{3}{c|}{\textbf{Vehicle (\textit{IoU=0.5})}} & \multicolumn{3}{c|}{\textbf{Pedestrian (\textit{IoU=0.25})}} & \multicolumn{3}{c}{\textbf{Cyclist (\textit{IoU=0.25})}} \\
        & Easy & Middle & Hard & Easy & Middle & Hard & Easy & Middle & Hard \\
        
        \Xhline{0.75pt}
        BEVHeight (ConvNeXt-B)
        & 78.08 & 65.99 & 66.07 & 41.76 & 40.84 & 40.03 & 58.76 & 60.69 & 60.76\\
        % \rowcolor{gray!15}
        BEVSpread (ConvNeXt-B)
        & \textbf{79.29} & \textbf{67.03} & \textbf{67.09} & \textbf{47.06} & \textbf{44.97} & \textbf{45.14} & \textbf{62.34} & \textbf{64.14} & \textbf{64.60} \\
        \textit{w.r.t. BEVHeight}
        &\textcolor{red}{+1.21}&\textcolor{red}{+1.04}&\textcolor{red}{+1.02}&\textcolor{red}{+5.30}&\textcolor{red}{+4.13}&\textcolor{red}{+5.11}&\textcolor{red}{+3.58}&\textcolor{red}{+3.45}&\textcolor{red}{+3.84}\\        
        \bottomrule
    \end{tabular}
% }
    }
    \vskip -0.3ex
\end{table*}

\vspace{0.5em} \noindent \textbf{Analysis on Different Backbones.}
We further compared BEVSpread with BEVHeight using different backbones. Results of ResNet-50/101 are listed in \cref{tab:dair} and \cref{tab:dair_spread}, and experiments for ConvNeXt-B can be found in \cref{tab:backbones}. Results show that stronger backbones lead to greater performance and our method can further improve it.

\section{Limitations and Analysis}
\label{sec:Limitations}
% We have illustrated the effectiveness of the proposed BEVSpread through experimental performance and visualization analysis. If a complete theoretical proof is given, the reliability of this method can be further improved. This issue is one of our future research directions.
% Besides, robustness experiments are conducted under extreme conditions. When the 2D-to-3D mapping is completely accurate, where height prediction uses GT, BEVSpread outperforms BEVHeight by a margin of (0.32, 4.29 and 0.96) AP in three categories. When the 2D-to-3D mapping is completely inaccurate, where height prediction uses Uniform distribution instead, BEVSpread outperforms BEVHeight by a margin of (0.73, 0.17 and 1.08) AP in three categories. We observe that, under extreme conditions, BEVSpread obtains a certain degree of performance improvement. However, the improvement is not as significant as in normal conditions.

\begin{table}[!t]
  \centering
    \resizebox{0.48 \textwidth}{!}{
    \begin{tabular}{l|cccc}
    \toprule
    \textbf{Method} & \textbf{Neighbors} & \textbf{Avg-AP} $\uparrow$ & \textbf{Latency-Total(ms)} $\downarrow$ & \textbf{Latency-Pooling(ms)} $\downarrow$ \\
    \midrule
    % \rowcolor{gray!15} \textbf{BEV GridSize=0.4} & & & & & \\
     BEVHeight (ResNet-101) & k = 1 & 56.69 &  74.3 &  5.5\\
     BEVSpread (ResNet-101) & k = 1 & 56.69(\textcolor{red}{+0.00}) &  69.8(\textcolor{red}{-6.1\%}) &  0.8\\
     \rowcolor{gray!24} BEVSpread (ResNet-101) & k = 2 & 58.68(\textcolor{red}{+1.99}) &  73.9(\textcolor{red}{-0.5\%}) &  4.9\\
     BEVSpread (ResNet-101) & k = 3 & 59.01(\textcolor{red}{+2.32}) &  76.6(\textcolor{green}{+3.1\%}) &  7.7\\
     BEVSpread (ResNet-101) & k = 6 & 59.83(\textcolor{red}{+3.14}) &  85.6(\textcolor{green}{+15.2\%}) &  15.3 \\
     \midrule
     BEVHeight (ResNet-50) & k = 1 & 55.90 & 61.4 &  5.5\\
     BEVSpread (ResNet-50) & k = 1 & 55.90(\textcolor{red}{+0.00}) & 57.1(\textcolor{red}{-7.0\%}) &  0.8\\
     \rowcolor{gray!24} BEVSpread (ResNet-50) & k = 2 & 58.12(\textcolor{red}{+2.22}) & 61.6(\textcolor{green}{+0.3\%}) &  4.9\\
     BEVSpread (ResNet-50) & k = 3 & 58.55(\textcolor{red}{+2.65}) & 64.2(\textcolor{green}{+4.6\%}) &  7.7\\
    \toprule
  \end{tabular}
  }
  % \vspace{-.2cm}
  \caption{Speed under different neighbor size $k$ on DAIR-V2X-I.}
  \vspace{-.5cm}
  \label{tab:speed}
  \vspace{-.3cm}
\end{table}

\vspace{-0.2cm} The proposed spread-voxel pooling brings a certain amount of calculation, resulting in an increase in latency.
While our approach is flexible to balance accuracy and speed by adjusting the spread scope, which is denoted as neighbor size $k$. As shown in Table \ref{tab:speed}, when $k$=2, BEVSpread still achieves significant improvement in Avg-AP without latency increase, benefiting from our CUDA optimization. Besides, the coordinates of these spread points are calculated online in this version. During the practical deployment phase, BEVSpread can use a preprocessing look-up table, akin to BEVPoolv2, for enhanced acceleration.

\section{Conclusion}
% \vspace{-.2cm}
\label{sec:Conclusion}
In this paper, we point out a approximation error in the current voxel pooling method. We proposed a novel voxel pooling strategy named BEVSpread to reduce this error. BEVSpread considers each frustum point as a source and spreads the image features to the surrounding BEV grids with adaptive weights. Additionally, a specific weight function is designed to dynamically control the decay speed based on distance and depth. Experiments in DAIR-V2X-I and Rope3D show that BEVSpread significantly improves the performance of existing frustum-based BEV methods.
\vspace{-.1em}\\

\noindent\textbf{Acknowledgements.} This work is supported in part by Zhejiang Provincial Natural Science Foundation of China under Grant LD24F020016, National Natural Science Foundation of China under Grant U20A20222, National Science Foundation for Distinguished Young Scholars under Grant 62225605, Zhejiang Key Research and Development Program under Grant 2023C03196, Baidu, SupreMind and The Ng Teng Fong Charitable Foundation in the form of ZJU-SUTD IDEA Grant, 188170-11102.
\clearpage
{
    \small
    \bibliographystyle{ieeenat_fullname}
    \bibliography{main}

\begin{thebibliography}{47}
\providecommand{\natexlab}[1]{#1}
\providecommand{\url}[1]{\texttt{#1}}
\expandafter\ifx\csname urlstyle\endcsname\relax
  \providecommand{\doi}[1]{doi: #1}\else
  \providecommand{\doi}{doi: \begingroup \urlstyle{rm}\Url}\fi

\bibitem[Creß et~al.(2022)Creß, Zimmer, Strand, Fortkord, Dai, Lakshminarasimhan, and Knoll]{A9-Dataset}
Christian Creß, Walter Zimmer, Leah Strand, Maximilian Fortkord, Siyi Dai, Venkatnarayanan Lakshminarasimhan, and Alois Knoll.
\newblock A9-dataset: Multi-sensor infrastructure-based dataset for mobility research.
\newblock In \emph{IV}, pages 965--970, 2022.

\bibitem[Fan et~al.(2022)Fan, Pang, Zhang, Wang, Zhao, Wang, Wang, and Zhang]{SST}
Lue Fan, Ziqi Pang, Tianyuan Zhang, Yu-Xiong Wang, Hang Zhao, Feng Wang, Naiyan Wang, and Zhaoxiang Zhang.
\newblock Embracing single stride 3d object detector with sparse transformer.
\newblock In \emph{CVPR}, pages 8458--8468, 2022.

\bibitem[Fan et~al.(2023)Fan, Wang, Huo, Wang, and Liu]{CBR}
Siqi Fan, Zhe Wang, Xiaoliang Huo, Yan Wang, and Jingjing Liu.
\newblock Calibration-free bev representation for infrastructure perception, 2023.
\newblock arXiv:2303.03583.

\bibitem[Geiger et~al.(2012)Geiger, Lenz, and Urtasun]{KITTI}
Andreas Geiger, Philip Lenz, and Raquel Urtasun.
\newblock Are we ready for autonomous driving? the kitti vision benchmark suite.
\newblock In \emph{ICCV}, pages 3354--3361, 2012.

\bibitem[He et~al.(2020)He, Zeng, Huang, Hua, and Zhang]{SA-SSD}
Chenhang He, Hui Zeng, Jianqiang Huang, Xian-Sheng Hua, and Lei Zhang.
\newblock Structure aware single-stage 3d object detection from point cloud.
\newblock In \emph{CVPR}, 2020.

\bibitem[He et~al.(2016)He, Zhang, Ren, and Sun]{Resnet}
Kaiming He, Xiangyu Zhang, Shaoqing Ren, and Jian Sun.
\newblock Deep residual learning for image recognition.
\newblock In \emph{CVPR}, 2016.

\bibitem[Hu et~al.(2022)Hu, Ding, Ge, Shao, Huang, Li, and Liu]{Afdetv2}
Yihan Hu, Zhuangzhuang Ding, Runzhou Ge, Wenxin Shao, Li Huang, Kun Li, and Qiang Liu.
\newblock Afdetv2: Rethinking the necessity of the second stage for object detection from point clouds.
\newblock In \emph{AAAI}, pages 969--979, 2022.

\bibitem[Huang and Huang(2022{\natexlab{a}})]{BEVDet4D}
Junjie Huang and Guan Huang.
\newblock Bevdet4d: Exploit temporal cues in multi-camera 3d object detection, 2022{\natexlab{a}}.
\newblock arXiv:2203.17054.

\bibitem[Huang and Huang(2022{\natexlab{b}})]{BEVPoolv2}
Junjie Huang and Guan Huang.
\newblock Bevpoolv2: A cutting-edge implementation of bevdet toward deployment, 2022{\natexlab{b}}.
\newblock arXiv:2211.17111.

\bibitem[Huang et~al.(2021)Huang, Huang, Zhu, Ye, and Du]{BEVDet}
Junjie Huang, Guan Huang, Zheng Zhu, Yun Ye, and Dalong Du.
\newblock Bevdet: High-performance multi-camera 3d object detection in bird-eye-view, 2021.
\newblock arXiv:2112.11790.

\bibitem[Jiang et~al.(2023)Jiang, Zhang, Miao, Zhu, Gao, Hu, and Jiang]{PolarFormer}
Yanqin Jiang, Li Zhang, Zhenwei Miao, Xiatian Zhu, Jin Gao, Weiming Hu, and Yu-Gang Jiang.
\newblock Polarformer: Multi-camera 3d object detection with polar transformer.
\newblock In \emph{AAAI}, pages 1042--1050, 2023.

\bibitem[Lang et~al.(2019)Lang, Vora, Caesar, Zhou, Yang, and Beijbom]{PointPillars}
Alex~H. Lang, Sourabh Vora, Holger Caesar, Lubing Zhou, Jiong Yang, and Oscar Beijbom.
\newblock Pointpillars: Fast encoders for object detection from point clouds.
\newblock In \emph{CVPR}, 2019.

\bibitem[Li et~al.(2022{\natexlab{a}})Li, Sima, Dai, Wang, Lu, Wang, Zeng, Li, Yang, Deng, Tian, Xie, Xie, Chen, Li, Li, Gao, Jia, Liu, Shi, Lin, and Qiao]{vision_BEV_Survey_2}
Hongyang Li, Chonghao Sima, Jifeng Dai, Wenhai Wang, Lewei Lu, Huijie Wang, Jia Zeng, Zhiqi Li, Jiazhi Yang, Hanming Deng, Hao Tian, Enze Xie, Jiangwei Xie, Li Chen, Tianyu Li, Yang Li, Yulu Gao, Xiaosong Jia, Si Liu, Jianping Shi, Dahua Lin, and Yu Qiao.
\newblock Delving into the devils of bird's-eye-view perception: A review, evaluation and recipe, 2022{\natexlab{a}}.
\newblock arXiv:2209.05324.

\bibitem[Li et~al.(2022{\natexlab{b}})Li, Bao, Ge, Yang, Sun, and Li]{Bevstereo}
Yinhao Li, Han Bao, Zheng Ge, Jinrong Yang, Jianjian Sun, and Zeming Li.
\newblock Bevstereo: Enhancing depth estimation in multi-view 3d object detection with dynamic temporal stereo, 2022{\natexlab{b}}.
\newblock arXiv:2209.10248.

\bibitem[Li et~al.(2022{\natexlab{c}})Li, Chen, Qi, Li, Sun, and Jia]{UVTR}
Yanwei Li, Yilun Chen, Xiaojuan Qi, Zeming Li, Jian Sun, and Jiaya Jia.
\newblock Unifying voxel-based representation with transformer for 3d object detection.
\newblock \emph{NeurIPS}, 35:\penalty0 18442--18455, 2022{\natexlab{c}}.

\bibitem[Li et~al.(2023)Li, Ge, Yu, Yang, Wang, Shi, Sun, and Li]{BEVDepth}
Yinhao Li, Zheng Ge, Guanyi Yu, Jinrong Yang, Zengran Wang, Yukang Shi, Jianjian Sun, and Zeming Li.
\newblock Bevdepth: Acquisition of reliable depth for multi-view 3d object detection.
\newblock In \emph{AAAI}, pages 1477--1485, 2023.

\bibitem[Li et~al.(2022{\natexlab{d}})Li, Wang, Li, Xie, Sima, Lu, Qiao, and Dai]{BEVFormer}
Zhiqi Li, Wenhai Wang, Hongyang Li, Enze Xie, Chonghao Sima, Tong Lu, Yu Qiao, and Jifeng Dai.
\newblock Bevformer: Learning bird’s-eye-view representation from multi-camera images via spatiotemporal transformers.
\newblock In \emph{ECCV}, pages 1--18, 2022{\natexlab{d}}.

\bibitem[Liang et~al.(2022)Liang, Xie, Yu, Xia, Lin, Wang, Tang, Wang, and Tang]{Bevfusion}
Tingting Liang, Hongwei Xie, Kaicheng Yu, Zhongyu Xia, Zhiwei Lin, Yongtao Wang, Tao Tang, Bing Wang, and Zhi Tang.
\newblock Bevfusion: A simple and robust lidar-camera fusion framework.
\newblock \emph{NeurIPS}, 35:\penalty0 10421--10434, 2022.

\bibitem[Liu et~al.(2022)Liu, Wang, Zhang, and Sun]{PETR}
Yingfei Liu, Tiancai Wang, Xiangyu Zhang, and Jian Sun.
\newblock Petr: Position embedding transformation for multi-view 3d object detection.
\newblock In \emph{ECCV}, pages 531--548, 2022.

\bibitem[Liu et~al.(2023)Liu, Yan, Jia, Li, Gao, Wang, and Zhang]{PETRv2}
Yingfei Liu, Junjie Yan, Fan Jia, Shuailin Li, Aqi Gao, Tiancai Wang, and Xiangyu Zhang.
\newblock Petrv2: A unified framework for 3d perception from multi-camera images.
\newblock In \emph{ICCV}, pages 3262--3272, 2023.

\bibitem[Loshchilov and Hutter(2019)]{AdamW}
Ilya Loshchilov and Frank Hutter.
\newblock Decoupled weight decay regularization, 2019.
\newblock arXiv:1711.05101.

\bibitem[Ma et~al.(2022)Ma, Wang, Bai, Yang, Hou, Wang, Qiao, Yang, Manocha, and Zhu]{vision_BEV_Survey_1}
Yuexin Ma, Tai Wang, Xuyang Bai, Huitong Yang, Yuenan Hou, Yaming Wang, Yu Qiao, Ruigang Yang, Dinesh Manocha, and Xinge Zhu.
\newblock Vision-centric bev perception: A survey, 2022.
\newblock arXiv:2208.02797.

\bibitem[Philion and Fidler(2020)]{LSS}
Jonah Philion and Sanja Fidler.
\newblock Lift, splat, shoot: Encoding images from arbitrary camera rigs by implicitly unprojecting to 3d.
\newblock In \emph{ECCV}, pages 194--210, 2020.

\bibitem[Qin et~al.(2023)Qin, Chen, Chen, Chen, and Li]{UniFusion}
Zequn Qin, Jingyu Chen, Chao Chen, Xiaozhi Chen, and Xi Li.
\newblock Unifusion: Unified multi-view fusion transformer for spatial-temporal representation in bird's-eye-view.
\newblock In \emph{ICCV}, pages 8690--8699, 2023.

\bibitem[Reading et~al.(2021)Reading, Harakeh, Chae, and Waslander]{CaDDN}
Cody Reading, Ali Harakeh, Julia Chae, and Steven~L. Waslander.
\newblock Categorical depth distribution network for monocular 3d object detection.
\newblock In \emph{CVPR}, pages 8555--8564, 2021.

\bibitem[Roddick et~al.(2018)Roddick, Kendall, and Cipolla]{OFT}
Thomas Roddick, Alex Kendall, and Roberto Cipolla.
\newblock Orthographic feature transform for monocular 3d object detection, 2018.
\newblock arXiv:1811.08188.

\bibitem[Roddick et~al.(2023)Roddick, Kendall, and Cipolla]{SA-BEV}
Thomas Roddick, Alex Kendall, and Roberto Cipolla.
\newblock Sa-bev: Generating semantic-aware bird's-eye-view feature for multi-view 3d object detection.
\newblock In \emph{ICCV}, 2023.

\bibitem[Rukhovich et~al.(2022)Rukhovich, Vorontsova, and Konushin]{ImVoxelNet}
Danila Rukhovich, Anna Vorontsova, and Anton Konushin.
\newblock Imvoxelnet: Image to voxels projection for monocular and multi-view general-purpose 3d object detection.
\newblock In \emph{WACV}, pages 2397--2406, 2022.

\bibitem[Shi et~al.(2023)Shi, Pang, Zhang, Yang, Wu, Ni, Lin, Stiefelhagen, and Wang]{CoBEV}
Hao Shi, Chengshan Pang, Jiaming Zhang, Kailun Yang, Yuhao Wu, Huajian Ni, Yining Lin, Rainer Stiefelhagen, and Kaiwei Wang.
\newblock Cobev: Elevating roadside 3d object detection with depth and height complementarity, 2023.
\newblock arXiv:2310.02815.

\bibitem[Shi et~al.(2020)Shi, Guo, Jiang, Wang, Shi, Wang, and Li]{PV-RCNN}
Shaoshuai Shi, Chaoxu Guo, Li Jiang, Zhe Wang, Jianping Shi, Xiaogang Wang, and Hongsheng Li.
\newblock Pv-rcnn: Point-voxel feature set abstraction for 3d object detection.
\newblock In \emph{CVPR}, 2020.

\bibitem[Simonelli et~al.(2019)Simonelli, Bulo, Porzi, Lopez-Antequera, and Kontschieder]{AP3D}
Andrea Simonelli, Samuel~Rota Bulo, Lorenzo Porzi, Manuel Lopez-Antequera, and Peter Kontschieder.
\newblock Disentangling monocular 3d object detection.
\newblock In \emph{ICCV}, 2019.

\bibitem[Sindagi et~al.(2019)Sindagi, Zhou, and Tuzel]{MVXNet}
Vishwanath~A. Sindagi, Yin Zhou, and Oncel Tuzel.
\newblock Mvx-net: Multimodal voxelnet for 3d object detection.
\newblock In \emph{ICRA}, pages 7276--7282, 2019.

\bibitem[Tan et~al.(2023{\natexlab{a}})Tan, Lyu, Li, Hu, Feng, Xu, and Yao]{AR2VP}
Jiayao Tan, Fan Lyu, Linyan Li, Fuyuan Hu, Tingliang Feng, Fenglei Xu, and Rui Yao.
\newblock Dynamic v2x autonomous perception from road-to-vehicle vision, 2023{\natexlab{a}}.
\newblock arXiv:2310.19113.

\bibitem[Tan et~al.(2023{\natexlab{b}})Tan, Lyu, Li, Hu, Feng, Xu, and Yao]{MonoGAE}
Jiayao Tan, Fan Lyu, Linyan Li, Fuyuan Hu, Tingliang Feng, Fenglei Xu, and Rui Yao.
\newblock Monogae: Roadside monocular 3d object detection with ground-aware embeddings, 2023{\natexlab{b}}.
\newblock arXiv:2310.00400.

\bibitem[Wang et~al.(2023{\natexlab{a}})Wang, Shi, Shi, Lei, Wang, He, Schiele, and Wang]{DSVT}
Haiyang Wang, Chen Shi, Shaoshuai Shi, Meng Lei, Sen Wang, Di He, Bernt Schiele, and Liwei Wang.
\newblock Dsvt: Dynamic sparse voxel transformer with rotated sets.
\newblock In \emph{CVPR}, pages 13520--13529, 2023{\natexlab{a}}.

\bibitem[Wang et~al.(2023{\natexlab{b}})Wang, Liu, Wang, Li, and Zhang]{StreamPETR}
Shihao Wang, Yingfei Liu, Tiancai Wang, Ying Li, and Xiangyu Zhang.
\newblock Exploring object-centric temporal modeling for efficient multi-view 3d object detection.
\newblock In \emph{ICCV}, 2023{\natexlab{b}}.

\bibitem[Wang et~al.(2022)Wang, Guizilini, Zhang, Wang, Zhao, and Solomon]{DETR3D}
Yue Wang, Vitor~Campagnolo Guizilini, Tianyuan Zhang, Yilun Wang, Hang Zhao, and Justin Solomon.
\newblock Detr3d: 3d object detection from multi-view images via 3d-to-2d queries.
\newblock In \emph{CoRL}, pages 180--191, 2022.

\bibitem[Wang et~al.(2023{\natexlab{c}})Wang, Chen, and Zhang]{FrustumFormer}
Yuqi Wang, Yuntao Chen, and Zhaoxiang Zhang.
\newblock Frustumformer: Adaptive instance-aware resampling for multi-view 3d detection.
\newblock In \emph{CVPR}, pages 5096--5105, 2023{\natexlab{c}}.

\bibitem[Wu et~al.(2023)Wu, Li, Qin, Zhao, and Li]{HeightFormer}
Yiming Wu, Ruixiang Li, Zequn Qin, Xinhai Zhao, and Xi Li.
\newblock Heightformer: Explicit height modeling without extra data for camera-only 3d object detection in bird's eye view, 2023.
\newblock arXiv:2307.13510.

\bibitem[Xie et~al.(2023)Xie, Xu, Rakotosaona, Rim, Tombari, Keutzer, Tomizuka, and Zhan]{SparseFusion}
Yichen Xie, Chenfeng Xu, Marie-Julie Rakotosaona, Patrick Rim, Federico Tombari, Kurt Keutzer, Masayoshi Tomizuka, and Wei Zhan.
\newblock Sparsefusion: Fusing multi-modal sparse representations for multi-sensor 3d object detection.
\newblock In \emph{ICCV}, 2023.

\bibitem[Yan et~al.(2023)Yan, Liu, Sun, Jia, Li, Wang, and Zhang]{CMT}
Junjie Yan, Yingfei Liu, Jianjian Sun, Fan Jia, Shuailin Li, Tiancai Wang, and Xiangyu Zhang.
\newblock Cross modal transformer: Towards fast and robust 3d object detection.
\newblock In \emph{ICCV}, pages 18268--18278, 2023.

\bibitem[Yan et~al.(2018)Yan, Mao, and Li]{Second}
Yan Yan, Yuxing Mao, and Bo Li.
\newblock Second: Sparsely embedded convolutional detection.
\newblock \emph{Sensors}, 18\penalty0 (10):\penalty0 3337, 2018.

\bibitem[Yang et~al.(2023{\natexlab{a}})Yang, Tang, Li, Chen, Yuan, Wang, Huang, Zhang, and Yu]{BEVHeight2}
Lei Yang, Tao Tang, Jun Li, Peng Chen, Kun Yuan, Li Wang, Yi Huang, Xinyu Zhang, and Kaicheng Yu.
\newblock Bevheight++: Toward robust visual centric 3d object detection, 2023{\natexlab{a}}.
\newblock arXiv:2309.16179.

\bibitem[Yang et~al.(2023{\natexlab{b}})Yang, Yu, Tang, Li, Yuan, Wang, Zhang, and Chen]{BEVHeight}
Lei Yang, Kaicheng Yu, Tao Tang, Jun Li, Kun Yuan, Li Wang, Xinyu Zhang, and Peng Chen.
\newblock Bevheight: A robust framework for vision-based roadside 3d object detection.
\newblock In \emph{CVPR}, pages 21611--21620, 2023{\natexlab{b}}.

\bibitem[Ye et~al.(2022)Ye, Shu, Li, Shi, Li, Wang, Tan, and Ding]{Rope3D}
Xiaoqing Ye, Mao Shu, Hanyu Li, Yifeng Shi, Yingying Li, Guangjie Wang, Xiao Tan, and Errui Ding.
\newblock Rope3d: The roadside perception dataset for autonomous driving and monocular 3d object detection task.
\newblock In \emph{CVPR}, pages 21341--21350, 2022.

\bibitem[Yu et~al.(2022)Yu, Luo, Shu, Huo, Yang, Shi, Guo, Li, Hu, Yuan, and Nie]{DAIR-V2X}
Haibao Yu, Yizhen Luo, Mao Shu, Yiyi Huo, Zebang Yang, Yifeng Shi, Zhenglong Guo, Hanyu Li, Xing Hu, Jirui Yuan, and Zaiqing Nie.
\newblock Dair-v2x: A large-scale dataset for vehicle-infrastructure cooperative 3d object detection.
\newblock In \emph{CVPR}, pages 21361--21370, 2022.

\bibitem[Yu et~al.(2023)Yu, Yang, Ruan, Yang, Tang, Gao, Hao, Shi, Pan, Sun, Song, Yuan, Luo, and Nie]{V2X-Seq}
Haibao Yu, Wenxian Yang, Hongzhi Ruan, Zhenwei Yang, Yingjuan Tang, Xu Gao, Xin Hao, Yifeng Shi, Yifeng Pan, Ning Sun, Juan Song, Jirui Yuan, Ping Luo, and Zaiqing Nie.
\newblock V2x-seq: A large-scale sequential dataset for vehicle-infrastructure cooperative perception and forecasting.
\newblock In \emph{CVPR}, pages 5486--5495, 2023.

\end{thebibliography}
}

% WARNING: do not forget to delete the supplementary pages from your submission 
\clearpage
\setcounter{page}{1}
\maketitlesupplementary
\setcounter{section}{0}
\renewcommand\thesection{\Alph{section}}
\section{Appendix}
\label{sec:Appendix}
\subsection{Potential Impacts}
We propose a novel spread voxel pooling approach, named BEVSpread, which is a plug-in and can enhance the performance of existing frustum-based BEV methods in roadsize perception. However, it may produce inaccurate predictions for autonomous vehicles, causing wrong decision-making and potential traffic accidents, and it may help tracking someone else, making privacy invasion happens. Compared with roadsize scenarios, vehicle-side perception is quite different, we think it is worth further research on how to apply spread voxel pooling on vehicle-side methods.

\subsection{Analysis on BEV grid size}
As shown in \cref{first_img}\textcolor{red}{a}, the predicted point is usually not located in a BEV grid center, previous work simply accumulates this point feature into its corresponding BEV grid, which causes a approximation error. As shown in \cref{tab:grid_size}, Augmenting the density of BEV grids can alleviate this error, and as BEV grid size decreases , the performance gradually improves. When the grid size is set to 0.2m, results of three categories are better than others.  However, augmenting the density of BEV grids results in a notable increase in computational workload and memory overhead, especially because of the long perception range in roadside scenarios. Therefore, when keep BEV gird a certain size, spread voxel pooling module can enhance the performance of existing frustum-based BEV methods without causing increased memory consumption.
% \subsection{Description of RegressNet}
% We have designed an intuitive experiment to verify the effectiveness of the proposed spread voxel pooling strategy. As shown in \cref{prove_framwork}, 3D points are randomly generated and projected onto the 16x16 BEV grids to obtain the BEV features based on voxel pooling and spread voxel pooling. RegressNet utilize U-Net encoder network to regress the accurate positions of these 3D points and MSE loss is used for network optimized. As shown in \cref{prove}, our spread voxel pooling recovers the random point position with 0.003 MSE loss when the neighbors number $\geq 3$, while the original voxel pooling obtains 0.095 MSE loss. 
\subsection{Analysis on Weight Function}
\begin{figure}[!htb]
    \centering
    \includegraphics[width=1.0\linewidth]{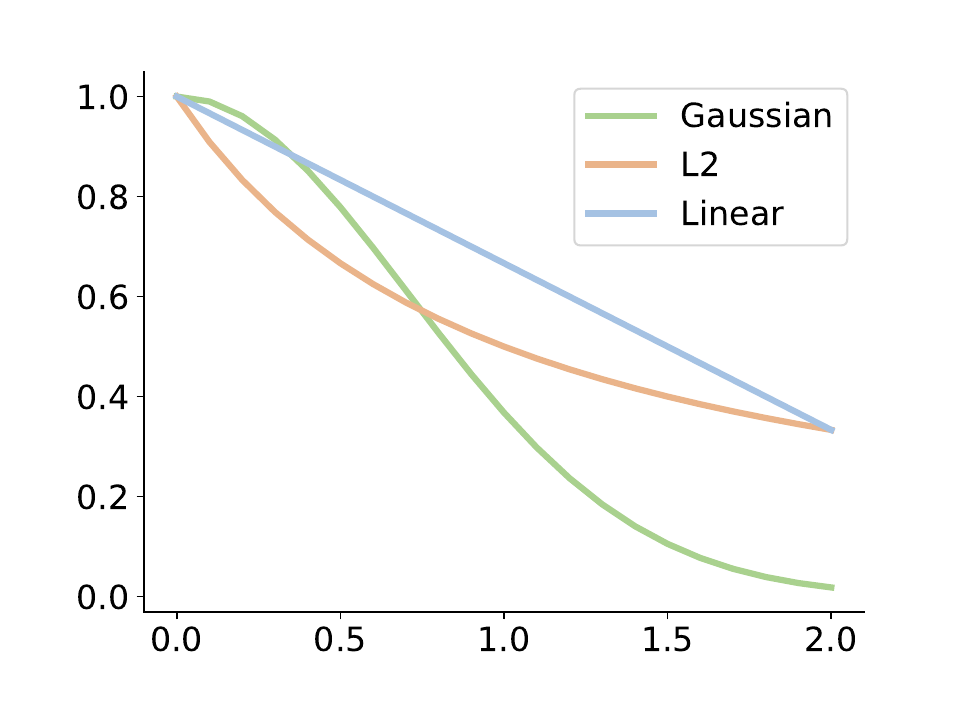}
    \caption{\textbf{Different weight functions.} We attempt a variety of functions, including L2, Linear and Gaussian distributions.}
    \label{func_weights}
\end{figure}

\begin{figure}[!htb]
    \centering
    \includegraphics[width=1.0\linewidth]{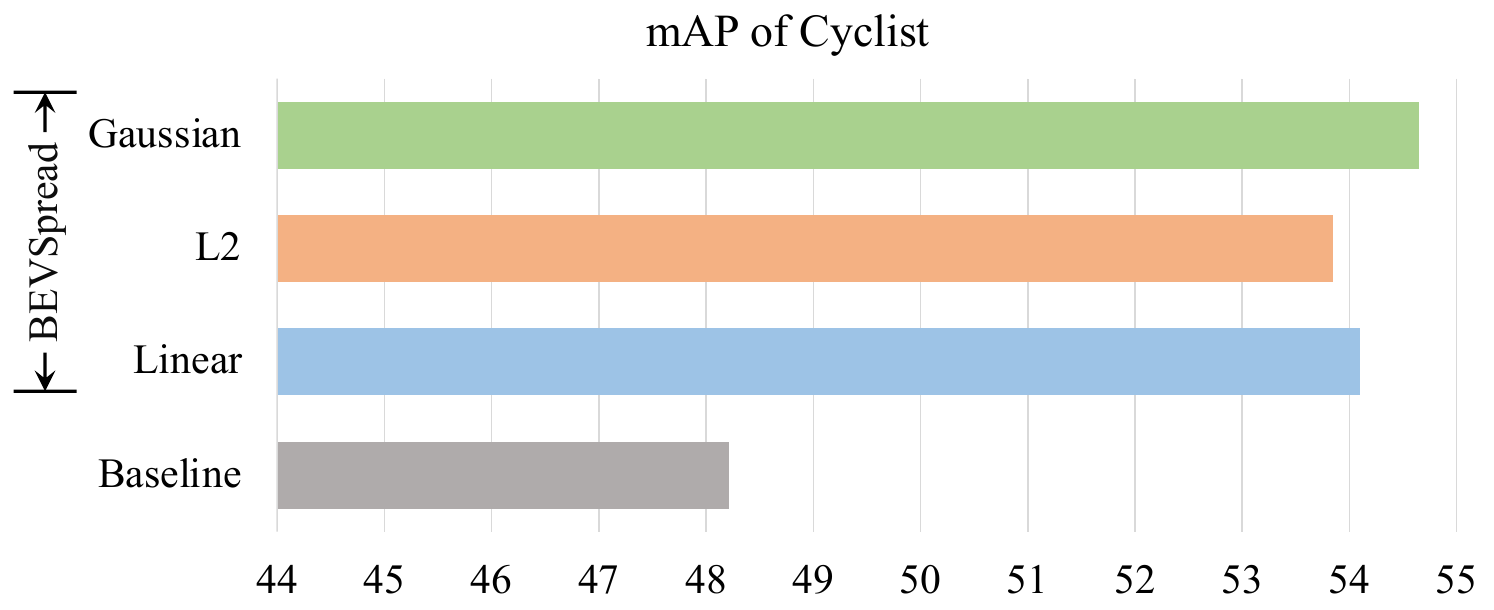}
    \caption{\textbf{Analysis on weight function.} BEVSpread is significantly better than baseline (BEVHeight \cite{BEVHeight}). And Gaussian function outperforms other counterparts.}
    \label{func_results}
\end{figure}

In order to achieve better performance, we attempt a variety of functions, including L2, Linear and Gaussian distributions, and their function curves are shown in \cref{func_weights}.  We compare the mAP of cyclist on DAIR-V2X-I \cite{DAIR-V2X} dataset with different weight functions, as shown in \cref{func_results}, BEVSpread is significantly better than baseline (BEVHeight \cite{BEVHeight}) and Gaussian function outperforms other counterparts. We believe that this is because Gaussian function's curve is smoother than others near original point, which makes it retaining more location information. And with distant increases, the weight decreases faster to 0, which prevents assigning information to wrong positions. The results of three categories with different weight functions on DAIR-V2X-I dataset can be found in \cref{tab:weight_decay}.

\begin{table*}[!t]
    \setlength{\tabcolsep}{4mm} % column spacing
    \tiny
    \renewcommand{\arraystretch}{1.2}
    \centering
    \caption{\textbf{Analysis on the BEV grid size.} Here we conduct experimets with different BEV grid size on DAIR-V2X-I dataset. BEVHeight \cite{BEVHeight} is used as base model, resnet-50 is used as image encoder, the BEV grid size is set to 0.8 meters, and the detection range is set to $0 {\sim} 100m$}
    \label{tab:grid_size}
    \vskip -2ex
    \resizebox{0.97\textwidth}{!}{
        
\resizebox{1.0\textwidth}{!}{
\setlength{\tabcolsep}{2mm}{ % 3mm
\begin{tabular}{c|ccc|ccc|ccc}
\hline
\multirow{2}{*}{\textbf{Grid Size}} &  \multicolumn{3}{c|}{\textbf{Vehicle (\textit{IoU=0.5})}} & \multicolumn{3}{c|}{\textbf{Pedestrian (\textit{IoU=0.25})}} & \multicolumn{3}{c}{\textbf{Cyclist (\textit{IoU=0.25})}} \\
& Easy & Middle & Hard & Easy & Middle & Hard & Easy & Middle & Hard \\
\hline
\hline
0.8m &76.59&64.69&64.76&27.08&25.79&25.29&49.39&52.30&52.84\\
0.4m &78.07&65.90&65.96&40.94&39.02&39.15&56.69&59.44&59.88\\
0.2m &\textbf{79.05}&\textbf{66.86}&\textbf{66.91}&\textbf{46.06}&\textbf{44.02}&\textbf{44.30}&\textbf{58.10}&\textbf{60.37}&\textbf{60.76}\\
\hline
\end{tabular}
}
}

    }
    \vskip -0.6ex
\end{table*}

\begin{table*}[!t]
    \setlength{\tabcolsep}{4mm} % column spacing
    \tiny
    \renewcommand{\arraystretch}{1.2}
    \centering
    \caption{\textbf{Analysis on Weight Function.} We attempt a variety of functions, including L2, Linear and Gaussian distributions. ResNet-50 is used as image encoder, the BEV grid size is set to 0.8 meters, and the detection range is set to $0 {\sim} 100m$}
    \label{tab:weight_decay}
    \vskip -2ex
    \resizebox{0.97\textwidth}{!}{
        \resizebox{1.0\textwidth}{!}{
\setlength{\tabcolsep}{2mm}{ % 3mm
\begin{tabular}{l|ccc|ccc|ccc}
\hline
\multirow{2}{*}{\textbf{Function}} &  \multicolumn{3}{c|}{\textbf{Vehicle (\textit{IoU=0.5})}} & \multicolumn{3}{c|}{\textbf{Pedestrian (\textit{IoU=0.25})}} & \multicolumn{3}{c}{\textbf{Cyclist (\textit{IoU=0.25})}} \\
& Easy & Middle & Hard & Easy & Middle & Hard & Easy & Middle & Hard \\
\hline
\hline
Base&76.24&64.54&64.13&26.47&25.79&25.72&48.55&48.21&47.96\\
\hline
Linear&77.44&65.43&65.51&31.31&29.84&30.08&52.60&54.10&54.67\\
L2&77.49&65.46&65.54&30.64&29.29&29.43&52.22&53.85&54.42\\
Gaussian &\textbf{77.67}&\textbf{65.61}&\textbf{65.69}&\textbf{31.34}&\textbf{29.94}&\textbf{30.08}&\textbf{53.53}&\textbf{54.65}&\textbf{55.17}\\
\hline
\end{tabular}
}
}
    }
    \vskip -0.6ex
\end{table*}

\begin{table*}[!t]
    \setlength{\tabcolsep}{4mm} % column spacing
    \tiny
    \renewcommand{\arraystretch}{1.2}
    \centering
    \caption{\textbf{Analysis on neighbors number.} For each neightbors number, we repeat 3 times. ResNet-101 is used as image encoder, the BEV grid size is set to 0.4 meters, and the detection range is set to $0 {\sim} 100m$}
    \label{tab:neighbors_number}
    \vskip -2ex
    \resizebox{0.97\textwidth}{!}{
        \resizebox{1.0\textwidth}{!}{
\setlength{\tabcolsep}{2mm}{ % 3mm
\begin{tabular}{c|ccc|ccc|ccc|c}
\hline
\multirow{2}{*}{\textbf{Neighbors Num}} &  \multicolumn{3}{c|}{\textbf{Vehicle (\textit{IoU=0.5})}} & \multicolumn{3}{c|}{\textbf{Pedestrian (\textit{IoU=0.25})}} & \multicolumn{3}{c|}{\textbf{Cyclist (\textit{IoU=0.25})}} & \multirow{2}{*}{\textbf{mAP}}\\
& Easy & Middle & Hard & Easy & Middle & Hard & Easy & Middle & Hard \\
\hline
\hline
\multirow{3}{*}{2} &78.57&66.34&66.43&45.80&43.74&43.89&58.20&61.09&61.48&\multirow{3}{*}{58.68}\\
&78.59&66.21&66.39&45.22&43.31&43.46&59.93&61.68&62.14\\
&78.73&66.47&66.54&44.38&42.54&42.76&62.89&63.54&63.96\\
\hline
\multirow{3}{*}{3}&79.00&66.69&66.76&45.86&43.90&44.01&62.26&62.88&63.19&\multirow{3}{*}{59.01}\\
&78.88&66.65&66.72&45.10&43.13&43.38&60.16&61.65&62.05\\
&78.92&66.68&66.76&45.40&43.52&43.66&61.21&62.22&62.56\\
\hline
\multirow{3}{*}{4}&79.01&66.77&66.83&44.56&42.63&42.76&63.16&63.40&63.73&\multirow{3}{*}{59.26}\\
&78.70&66.47&66.53&45.67&43.63&43.84&62.65&63.03&63.40\\
&78.78&66.63&66.72&45.25&43.30&43.47&62.82&63.17&63.26\\
\hline
\multirow{3}{*}{5}&78.60&66.38&66.44&45.21&43.14&43.36&62.76&63.41&63.78&\multirow{3}{*}{59.65}\\
&79.05&66.74&66.80&46.70&44.61&44.89&\textbf{63.55}&\textbf{63.87}&\textbf{64.21}\\
&78.80&66.65&66.59&46.05&44.00&44.30&63.07&63.70&63.97\\
\hline
\multirow{3}{*}{6}
&78.61&66.41&66.46&\textbf{47.05}&\textbf{45.05}&\textbf{45.29}&62.05&62.73&63.18&\multirow{3}{*}{\textbf{59.71}}\\
&78.80&66.56&66.53&46.53&44.39&44.70&62.53&63.26&63.67\\
&\textbf{79.07}&\textbf{66.82}&\textbf{66.88}&46.54&44.51&44.71&62.64&63.50&63.75\\
\hline
\end{tabular}
}
}
    }
    \vskip -0.6ex
\end{table*}

\begin{table*}[!t]
    \setlength{\tabcolsep}{4mm} % column spacing
    \tiny
    \renewcommand{\arraystretch}{1.1}
    \centering
    \caption{\textbf{Ablation study of spread voxel pooling on the Rope3D \cite{Rope3D}.}  ResNet-50 is used as image encoder, the BEV grid size is set to 0.8 meters, and the detection range is set to $0 {\sim} 100m$, and $\text{top-}k$ ($k$=2)  nearest BEV grid centers are selected as neighbors.}
    \label{tab:rope3d_spread}
    \vskip -2ex
    \resizebox{1.0\textwidth}{!}{
        \resizebox{1.0\textwidth}{!}{
\setlength{\tabcolsep}{2mm}{ % 3mm
\begin{tabular}{l|ccc|ccc|ccc}
\hline
\multirow{2}{*}{\textbf{Method}} &  \multicolumn{3}{c|}{\textbf{Vehicle (\textit{IoU=0.5})}} & \multicolumn{3}{c|}{\textbf{Pedestrian (\textit{IoU=0.25})}} & \multicolumn{3}{c}{\textbf{Cyclist (\textit{IoU=0.25})}} \\
& Easy & Middle & Hard & Easy & Middle & Hard & Easy & Middle & Hard \\
\hline
\hline
BEVDepth~\cite{BEVDepth} 
&75.90&65.14&65.10&16.82&16.14&16.31&52.53&51.70&49.81\\
+ spread voxel pooling 
&\textbf{79.27}&\textbf{68.19}&\textbf{68.17}&\textbf{21.97}&\textbf{21.09}&\textbf{21.19}&\textbf{54.95}&\textbf{54.20}&\textbf{54.13}\\
\textit{w.r.t. BEVDepth}
&\textcolor{red}{+3.37}&\textcolor{red}{+3.05}&\textcolor{red}{+3.07}&\textcolor{red}{+5.15}&\textcolor{red}{+4.95}&\textcolor{red}{+4.88}&\textcolor{red}{+2.43}&\textcolor{red}{+2.49}&\textcolor{red}{+4.32}\\
\hline
BEVHeight~\cite{BEVHeight}
&76.42&67.24&67.07&21.57&19.79&19.98&56.57&54.80&54.68\\
+ spread voxel pooling 
&\textbf{80.16}&\textbf{70.79}&\textbf{70.72}&\textbf{23.7}5&\textbf{21.70}&\textbf{21.00}&\textbf{59.34}&\textbf{57.34}&\textbf{57.23}\\
\textit{w.r.t. BEVHeight}
&\textcolor{red}{+3.74}&\textcolor{red}{+3.54}&\textcolor{red}{+3.66}&\textcolor{red}{+2.18}&\textcolor{red}{+1.90}&\textcolor{red}{+1.03}&\textcolor{red}{+2.77}&\textcolor{red}{+2.54}&\textcolor{red}{+2.56}\\
\hline
\end{tabular}
}
}
    }
    \vskip -2ex
\end{table*}

\begin{figure}[!htb]
    \centering
    \includegraphics[width=0.88\linewidth]{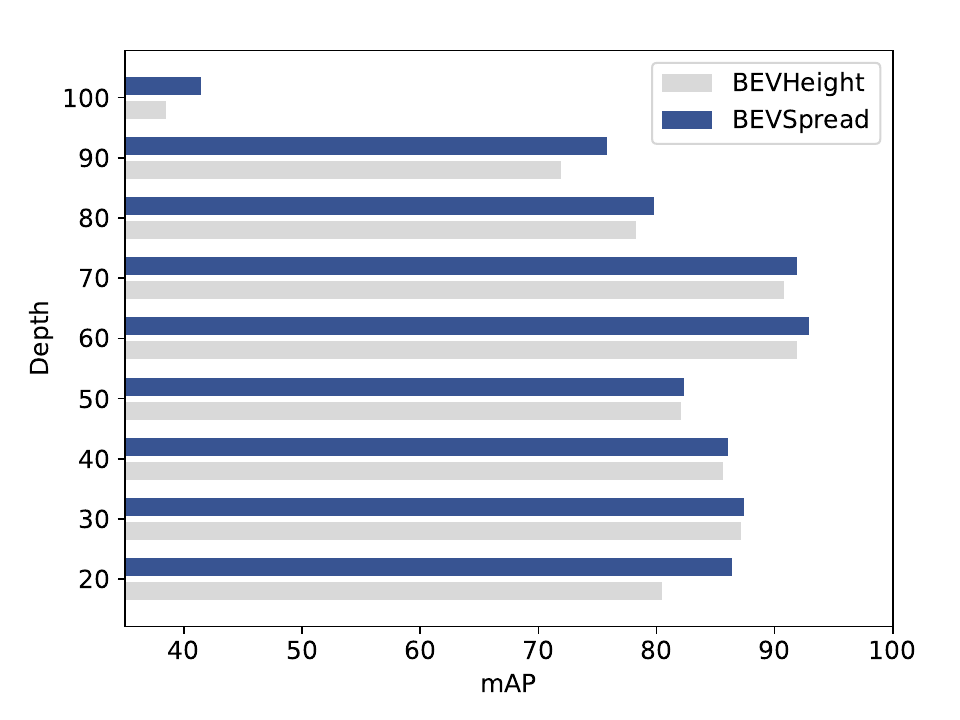}
    \caption{Range-wise evaluation on the DAIR-V2X-I validation set. Metric is $\text{AP}_{\text{3D}|\text{R40}}$ of the \textbf{Vehicle} category under moderate setting. The sample interval is 10m, e.g., the value at vertical axis 50 indicates the overall performance of all samples between 45m and 55m.}
    \label{range_wise_car}
\end{figure}

\subsection{Results on Neighbors Number}
To investigate the effect of the hyperparameter neighbors number on the performance of BEVSpread, We repect 3 times for each neighbors number k selection. \cref{neighbor} shows how the mAP of three categoties changes with neighbors number k. The light-blue area indicates the error range and it can be observed that the performance of $k \geq 2$ is significantly better than $k=1$ (baseline). As $k$ increases, the performance gradually improves and becomes stable. The specific experiment results are presented in \cref{tab:neighbors_number}.

\subsection{Ablation Study on Rope3D}
The proposed spread voxel pooling strategy, as a plug-in, can significantly improve the performance of existing frustum-based BEV methods. We further conduct ablation study on Rope3D dataset. As shown in \cref{tab:rope3d_spread}, after being deployed to BEVDepth \cite{BEVDepth}, the detection performance has been significantly improved by a margin of (3.16, 4.99 and 3.08) AP in three categories. After being deployed to BEVHeight \cite{BEVHeight}, the detection performance has been improved by a margin of (3.65, 1.70 and 2.62) AP in three categories.

\subsection{Robust of BEVSpread}
In real-world scenarios, roadside cameras are mounted on poles a few meters above the ground, and are often subjected to variations in extrinsic parameters caused by factors such as wind, vibrations, human adjustments, and other environmental conditions. Additionally, the intrinsic parameters also change between different cameras. So we investigate the robustness of BEVHeight and BEVSpread in the context of fluctuations in camera parameters. We introduce offset noise with a $N(0,1.67)$ distribution to $roll$ and $pitch$ angles associated with the extrinsic matrix. For the camera $focal$ length, we introduce scale noise, with the scaling coefficient following a $N(1,0.2)$ distribution. As shown in \cref{tab:robust}, BEVSpread maintains the best accuracy across all test-time scenarios involving noisy camera parameters. When only the $pitch$ angle is disturbed, BEVSpread exhibits significantly enhanced robustness compared to BEVHeight, with an improvement of (3.29, 8.75 and 7.71) AP in three categories.  These results reveal BEVSpread’s excellent robustness and resistance to interference.

\begin{figure}[!htb]
    \centering
    \includegraphics[width=0.88\linewidth]{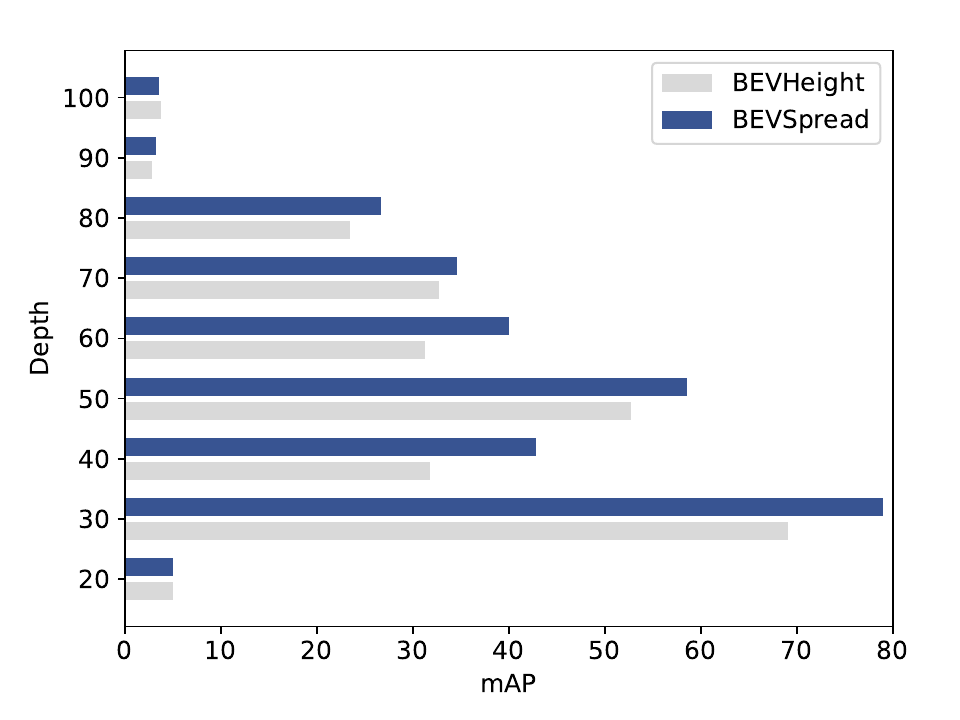}
    \caption{Range-wise evaluation on the DAIR-V2X-I validation set. Metric is $\text{AP}_{\text{3D}|\text{R40}}$ of the \textbf{Pedestrian} category under moderate setting. The sample interval is 10m, e.g., the value at vertical axis 50 indicates the overall performance of all samples between 45m and 55m.}
    \label{range_wise_ped}
\end{figure}

\subsection{Range-wise Evaluation}
We present the accuracy distributions of BEVHeight \cite{BEVHeight} and the proposed BEVSpread within different range intervals. As shown in \cref{range_wise_car}, we can observe that BEVSpread exhibit a notable advantage on vehicle category in long-range scenarios, particularly at distances of more than 50 meters. We believe that this advantage stems from the fact
that BEVSpread assigns larger weights to the surrounding BEV grids for distant targets, which results in distant objects containing more image features, leading to a better performance in long-range scenarios. As shown in \cref{range_wise_ped}, BEVSpread outperforms BEVHeight at all distances on pedestrian category. We believe that this is because BEVSpread can reduce the approximation error in voxel pooling, which significantly influences the detection of small scale objects like pedestrian.

\subsection{Customized CUDA Parallel Acceleration}
We refined the cuda acceleration operator of origin voxel pooling. Except for parallel processing of all points, we further parallelize the addition of the 80 channels of context features, which makes voxel pooling faster. BEVHeight takes 74.3ms for one inference and under same configuration, BEVSpread takes 69.8ms when neighbors number is set to 1 and takes 73.9ms when neighbors number is set to 2. As shown in \cref{neighbor}, when neightbors number $k = 2$, the performance of BEVSpread is better than the baseline. So we can make a balance between accuracy and inference time, and BEVSpread can achieve comparable inference time as BEVHeight while significantly improves its performance.

\subsection{More Visualizations}
\begin{figure*}[!htb]
    \centering
    \includegraphics[width=1\linewidth]{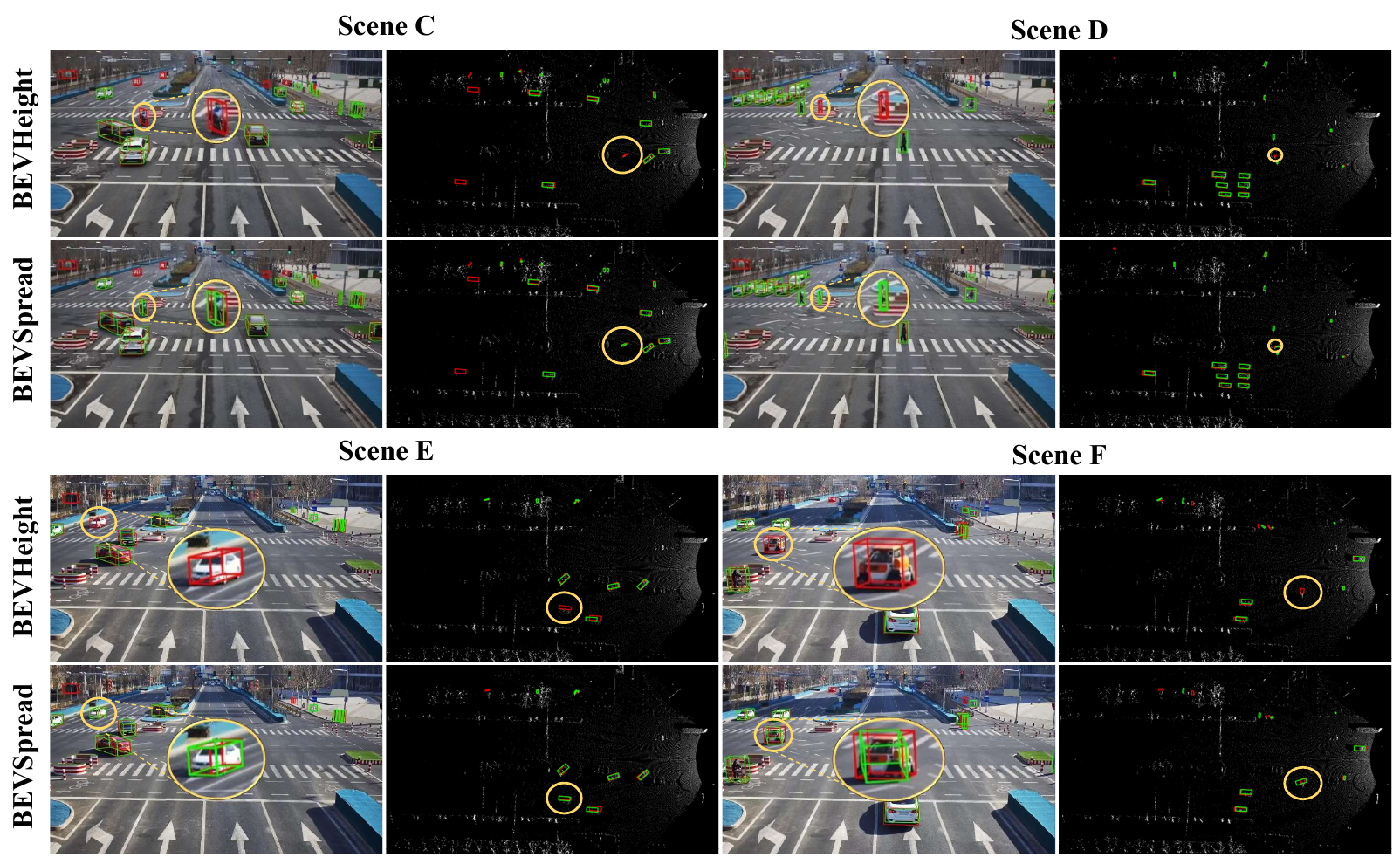}
    \caption{More Visualization results of BEVHeight and proposed BEVSpread in image and BEV view.}
    \label{more_vis}
    % \vskip 4ex
\end{figure*}

In \cref{more_vis}, we present more visuaization results on the DAIR-V2X-I \cite{DAIR-V2X} dataset. We can see that BEVSpread has advantages in detection of long-range objects and small scale objects.

\begin{table*}[!t]
    \setlength{\tabcolsep}{2mm} % column spacing
    \renewcommand{\arraystretch}{1.1}
    \centering
    \caption{\textbf{Robustness analysis on the DAIR-V2X-I validation set.} Three disturbed factors of roadside cameras are investigated, including focal length, roll angle, and pitch angle.}
    \label{tab:disturb}
    \vskip -1ex
    \resizebox{1.0\textwidth}{!}{
        \resizebox{0.9\textwidth}{!}{
\setlength{\tabcolsep}{2mm}{ % 4mm
\begin{tabular}{lccc|ccc|ccc|ccc}
\hline
\multirow{2}{*}{\textbf{Method}} & \multicolumn{3}{c|}{\textbf{Disturbed}} & \multicolumn{3}{c|}{\textbf{Vehicle $(IoU=0.5)$}} & \multicolumn{3}{c|}{\textbf{Pedestrian $(IoU=0.25)$}} & \multicolumn{3}{c}{\textbf{Cyclist $(IoU=0.25)$}} \\
& focal & roll & pitch & Easy & Middle & Hard & Easy & Middle & Hard & Easy & Middle & Hard \\
\hline
\hline

\multirow{6}{*}{BEVDepth\cite{BEVDepth}} & - & - & - & 75.31 & 65.24 & 65.32 & 32.68 & 31.01 & 31.33 & 46.96 & 50.88 & 51.44 \\
 & \checkmark & - & - & 72.17 & 60.19 & 60.20 & 25.75 & 25.16 & 24.35 & 40.65 & 47.09 & 47.21 \\
 & - & \checkmark & - & 74.78 & 62.72 & 62.81 & 30.80 & 30.20 & 30.43 & 45.58 & 50.07 & 50.72 \\
 & - & - & \checkmark & 74.83 & 62.76 & 62.85 & 30.21 & 28.62 & 28.91 & 46.07 & 50.15 & 50.85 \\
 & - & \checkmark & \checkmark & 74.62 & 62.57 & 62.66 & 30.38 & 28.87 & 29.13 & 45.96 & 50.15 & 50.79 \\
 \rowcolor{gray!20}
 & \checkmark & \checkmark & \checkmark & 71.91 & 59.94 & 59.96 & 26.61 & 25.18 & 25.22 & 39.79 & 46.11 & 46.13 \\

\hline

\multirow{6}{*}{BEVHeight\cite{BEVHeight}} & - & - & - & 78.08 & 65.97 & 66.04 & 40.01 & 38.21 & 38.38 & 58.01 & 60.46 & 60.95 \\
 & \checkmark & - & - & 72.30 & 60.45 & 60.47 & 32.18 & 30.65 & 29.65 & 50.06 & 55.04 & 55.14 \\
 & - & \checkmark & - & 77.65 & 65.57 & 65.65 & 38.38 & 36.60 & 36.72 & 56.15 & 59.11 & 59.52 \\
 & - & - & \checkmark & 75.37 & 63.31 & 63.38 & 33.13 & 31.47 & 31.63 & 52.88 & 56.07 & 56.44 \\
 & - & \checkmark & \checkmark & 75.06 & 63.08 & 63.16 & 33.67 & 31.19 & 31.30 & 51.65 & 54.93 & 56.83 \\
 \rowcolor{gray!20}
 & \checkmark & \checkmark & \checkmark & 71.71 & 59.92 & 59.96 & 27.81 & 26.43 & 26.36 & 47.42 & 51.19 & 51.26 \\

\hline

\multirow{7}{*}{BEVSpread (Ours)} 
 & - & - & -                        & 79.15 & 66.86 & 66.92 & 46.64 & 44.61 & 44.73 & 63.15 & 63.55 & 63.94 \\
 & \checkmark & - & -               & 75.41 & 63.37 & 63.40 & 35.09 & 35.09 & 33.33 & 53.61 & 56.98 & 56.90 \\
 & - & \checkmark & -               & 78.44 & 66.20 & 66.29 & 42.19 & 39.27 & 40.33 & 59.93 & 62.75 & 63.15 \\
 & - & - & \checkmark               & 78.84 & 66.51 & 66.59 & 42.03 & 40.14 & 40.30 & 61.22 & 63.47 & 63.84 \\
 & - & \checkmark & \checkmark      & 77.96 & 65.77 & 65.87 & 39.69 & 37.87 & 37.96 & 59.90 & 62.51 & 62.86 \\
 \rowcolor{gray!20}
 & \checkmark & \checkmark & \checkmark & \textbf{72.66} & \textbf{60.70} & \textbf{60.70} & \textbf{32.41} & \textbf{30.73} & \textbf{30.71} & \textbf{51.52} & \textbf{56.69} & \textbf{56.67} \\
 \rowcolor{gray!20}
 & \multicolumn{3}{c|}{\textit{w.r.t. BEVHeight}} & \textcolor{red}{+0.95} & \textcolor{red}{+0.78} & \textcolor{red}{+0.78} & \textcolor{red}{+4.60} & \textcolor{red}{+4.30} & \textcolor{red}{+4.35} & \textcolor{red}{+4.10} & \textcolor{red}{+5.50} & \textcolor{red}{+5.41} \\
\hline
\end{tabular}
}
}
    }
    \vskip -0.5ex
\end{table*}

% % 
% Having the supplementary compiled together with the main paper means that:
% % 
% \begin{itemize}
% \item The supplementary can back-reference sections of the main paper, for example, we can refer to \cref{sec:intro};
% \item The main paper can forward reference sub-sections within the supplementary explicitly (e.g. referring to a particular experiment); 
% \item When submitted to arXiv, the supplementary will already included at the end of the paper.
% \end{itemize}
% % 
% To split the supplementary pages from the main paper, you can use \href{https://support.apple.com/en-ca/guide/preview/prvw11793/mac#:~:text=Delete%20a%20page%20from%20a,or%20choose%20Edit%20%3E%20Delete).}{Preview (on macOS)}, \href{https://www.adobe.com/acrobat/how-to/delete-pages-from-pdf.html#:~:text=Choose%20%E2%80%9CTools%E2%80%9D%20%3E%20%E2%80%9COrganize,or%20pages%20from%20the%20file.}{Adobe Acrobat} (on all OSs), as well as \href{https://superuser.com/questions/517986/is-it-possible-to-delete-some-pages-of-a-pdf-document}{command line tools}.

\end{document}